\documentclass{article}

\usepackage[main, final]{neurips_2026}

\usepackage[utf8]{inputenc}
\usepackage[T1]{fontenc}
\usepackage{hyperref}
\usepackage{float}
\usepackage{url}
\usepackage{booktabs}
\usepackage{amsfonts}
\usepackage{amsmath}
\usepackage{amssymb}
\usepackage{nicefrac}
\usepackage{multirow}
\usepackage{microtype}
\usepackage{xcolor}
\usepackage{graphicx}
\usepackage{subcaption}
\usepackage{natbib}
\setcitestyle{numbers,square,sort&compress}
\usepackage{comment}
\usepackage{wrapfig}
\usepackage{tikz}
\usetikzlibrary{arrows.meta, positioning, calc, fit}

\title{TIDES: Implicit Time-Awareness in Selective State Space Models}

\author{%
  Taylan Soydan \\
  AIMM, ETH Zürich \\
  Zurich, Switzerland \\
  \And
  Miguel A. Bessa \\
  School of Engineering \\
  Brown University \\
  Providence, RI, USA \\
  \And
  Dirk Mohr \\
  AIMM, ETH Zürich \\
  Zurich, Switzerland \\
  \And
  Rui Barreira\thanks{Corresponding author:
  \texttt{rbarreira@ethz.ch}.} \\
  Inspire AG\\
  AIMM, ETH Zürich \\
  Zurich, Switzerland \\
}

\begin{document}

\maketitle

\begin{abstract}
Selective state space models (SSMs), such as Mamba, achieve strong per-token expressivity by making the time discretization step $\Tilde{\Delta}$ a learned function of the input. However, in doing so, $\Tilde{\Delta}$ no longer equals the physical time gap $\Delta$ between consecutive observations, limiting the ability of these models to handle irregular time series. Continuous time SSMs, such as S5, keep $\Tilde{\Delta}\equiv\Delta$ and therefore handle irregular timestamps natively, but their dynamics remain linear time invariant (LTI), limiting per token expressivity. We propose \textbf{TIDES}, a selective SSM variant that reconciles selective and continuous architectures by moving input dependence off the step size and onto the diagonal state matrix. As a result, $\Tilde{\Delta}\equiv\Delta$ as in S5, allowing the model to handle irregular timestamps natively without sacrificing the per-token expressivity that makes selective SSMs effective. We show this on a novel \emph{Fading Flash} experimental benchmark, a compact controlled diagnostic for sequence models that jointly tests input dependence and extrapolation to out-of-distribution $\Delta$ values, and isolates the distinct failure modes of current state-of-the-art architectures that TIDES avoids by construction. On large-scale benchmarks, TIDES sets the new best average rank on UEA time series classification and the Physiome ODE regression benchmark, and matches or exceeds the reference baseline model on 6 of 8 natively irregular datasets from astronomy, agriculture, neuromorphic sensing, and climate events.
Code available at: \url{https://github.com/TaylanSoydan/TIDES}.
\end{abstract}

\section{Introduction}
\label{sec:intro}

Real world sequential data is rarely sampled on a regular grid. Clinical observations arrive whenever a measurement happens to be ordered. Wearable devices often sample when they wake up, when a threshold is crossed, or whenever connectivity permits. Financial events occur at intervals that span many orders of magnitude. Scientific and industrial instruments record on whatever cadence their availability allows, as in seismology or astronomical surveys, where the phenomena of interest unfold outside the experimenter's control. Asynchronous systems with many channels add a further source of irregularity. In all of these settings, the elapsed time between observations carries information about the underlying continuous process, and a robust model must handle it correctly.

Sequence models that assume a uniform step between tokens must either resample these sequences (discarding information about \emph{when} events happened) or be augmented with explicit time-handling machinery. The latter approach dominates in the irregular-time-series literature: methods such as mTAN \citep{mtan_shukla2021multi}, CRU \citep{cru_schirmer2022modeling}, GRU-D \citep{grud_che2018recurrent}, and Latent ODE \citep{rubanova2019latent} encode time as an auxiliary feature, parameterize an attention bias by elapsed time, or solve an ordinary differential equation (ODE) between observations.

In parallel, two lineages of state space models (SSMs) have emerged as strong general sequence models. Continuous-time diagonal SSMs, such as S5~\citep{smith2023s5}, discretize a linear system $\dot{x} = Ax + Bu$ using the physical interval between consecutive data observations, $\Delta$, as their integration step, $\Tilde{\Delta}$, i.e., $\Tilde{\Delta}=\Delta$. The diagonal entries of $A$ are complex-valued: the real part controls how fast each state component decays, the imaginary part sets its oscillation frequency. Because $\Tilde{\Delta}$ is equivalent to the physical sampling time, these architectures handle irregularly sampled sequences naturally: changing the spacing between observations reshapes the discretized dynamics in the way the underlying continuous system would respond. The tradeoff is that the resulting discrete evolution is time-invariant, so identical inputs produce identical state updates regardless of context.

Selective SSMs such as Mamba~\citep{gu2023mamba} and its successors~\citep{dao2024transformers, mamba3} removed this rigidity by making $\Tilde{\Delta}$ and the input and output projections functions of the input. These models can contract or expand their effective time constant token-by-token, which is the source of their strong language modeling performance. But this comes at a cost: $\Tilde{\Delta}$ is no longer a physical sampling interval, i.e., $\Tilde{\Delta}\neq \Delta$, but a learned, content-dependent gate. A real, irregular timestamp has nowhere to enter the model unless it is appended to the input as a feature, in which case the model must learn the relationship between this feature and its internal gate from data, a relationship that, as we show, does not extrapolate beyond the training distribution.

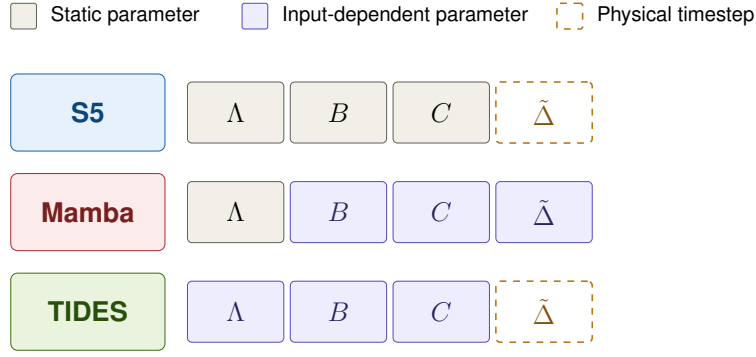
\begin{figure}[t]
\centering

\definecolor{staticFill}{HTML}{F1EFE8}
\definecolor{staticStroke}{HTML}{5F5E5A}
\definecolor{idFill}{HTML}{EEEDFE}
\definecolor{idStroke}{HTML}{534AB7}
\definecolor{idText}{HTML}{26215C}
\definecolor{physStroke}{HTML}{BA7517}
\definecolor{physText}{HTML}{854F0B}
\definecolor{s5Fill}{HTML}{E6F1FB}
\definecolor{s5Stroke}{HTML}{185FA5}
\definecolor{s5Text}{HTML}{0C447C}
\definecolor{mambaFill}{HTML}{FCEBEB}
\definecolor{mambaStroke}{HTML}{A32D2D}
\definecolor{mambaText}{HTML}{791F1F}
\definecolor{tidesFill}{HTML}{EAF3DE}
\definecolor{tidesStroke}{HTML}{3B6D11}
\definecolor{tidesText}{HTML}{27500A}

\scalebox{0.85}{%
\begin{tikzpicture}[
    font=\sffamily,
    pbox/.style={
        rounded corners=2pt, draw, line width=0.4pt,
        minimum width=1.5cm, minimum height=0.95cm,
        anchor=center, inner sep=2pt
    },
    mbox/.style={
        rounded corners=3pt, draw, line width=0.4pt,
        text width=3.2cm, align=center,
        minimum height=0.95cm,
        anchor=center, inner sep=1pt,
        font=\sffamily\bfseries\large
    },
    legchip/.style={
        rounded corners=1pt, line width=0.4pt,
        minimum width=0.4cm, minimum height=0.4cm
    }
]

\node[legchip, draw=staticStroke, fill=staticFill] at (0.2, 4.6) {};
\node[anchor=west, font=\sffamily\small] at (0.5, 4.6) {Static parameter};

\node[legchip, draw=idStroke, fill=idFill] at (3.8, 4.6) {};
\node[anchor=west, font=\sffamily\small] at (4.1, 4.6) {Input-dependent parameter};

\node[legchip, draw=physStroke, dashed, line width=0.7pt] at (8.7, 4.6) {};
\node[anchor=west, font=\sffamily\small] at (9.0, 4.6) {Physical timestep};

\node[mbox, draw=s5Stroke, fill=s5Fill, text=s5Text] at (1.2, 3.1) {S5};

\node[pbox, draw=staticStroke, fill=staticFill] at (4.2, 3.1) {\large $\Lambda$};
\node[pbox, draw=staticStroke, fill=staticFill] at (5.8, 3.1) {\large $B$};
\node[pbox, draw=staticStroke, fill=staticFill] at (7.4, 3.1) {\large $C$};

\node[pbox, draw=physStroke, dashed, line width=0.7pt, text=physText] at (9.0, 3.1) {\large $\Tilde{\Delta}$};

\node[mbox, draw=mambaStroke, fill=mambaFill, text=mambaText] at (1.2, 1.55) {Mamba};

\node[pbox, draw=staticStroke, fill=staticFill] at (4.2, 1.55) {\large $\Lambda$};
\node[pbox, draw=idStroke, fill=idFill, text=idText] at (5.8, 1.55) {\large $B$};
\node[pbox, draw=idStroke, fill=idFill, text=idText] at (7.4, 1.55) {\large $C$};
\node[pbox, draw=idStroke, fill=idFill, text=idText] at (9.0, 1.55) {\large $\Tilde{\Delta}$};

\node[mbox, draw=tidesStroke, fill=tidesFill, text=tidesText] at (1.2, 0) {TIDES};

\node[pbox, draw=idStroke, fill=idFill, text=idText] at (4.2, 0) {\large $\Lambda$};
\node[pbox, draw=idStroke, fill=idFill, text=idText] at (5.8, 0) {\large $B$};
\node[pbox, draw=idStroke, fill=idFill, text=idText] at (7.4, 0) {\large $C$};

\node[pbox, draw=physStroke, dashed, line width=0.7pt, text=physText] at (9.0, 0) {\large $\Tilde{\Delta}$};

\end{tikzpicture}
}

\caption{Where input-dependence lives in each architecture. S5 keeps all parameters static. Mamba makes $B$, $C$, and $\Tilde{\Delta}$ input-dependent, which collapses $\Tilde{\Delta}$ from a physical sampling interval into a learned function of the input. TIDES instead places input-dependence on $\Lambda$, $B$, and $C$, while leaving $\Tilde{\Delta}\equiv\Delta$ as the physical timestep, recovering selectivity without sacrificing irregular-time semantics.}
\label{fig:tides_state}
\end{figure}

\paragraph{Our approach.} Mamba and S5 disagree on the meaning of $\Tilde{\Delta}$. We propose a third design that avoids this tradeoff by routing input-dependence through the state matrix instead (Fig.~\ref{fig:tides_state}). Concretely, we make the per-component decay rate and the input and output projections functions of the input, while keeping the oscillation frequency static and leaving $\Tilde{\Delta}$ as a physical timestep quantity rather than a learned gate. Under irregular sampling, $\Tilde{\Delta}$ still varies from step to step, but it is set by the observation timestamps, not produced by a function of the input. Crucially, selectivity now lives in the continuous-time generator rather than in the discretization step, so expressivity is preserved without collapsing $\Tilde{\Delta}$. We call the resulting property \emph{selective implicit time-awareness}: the model's response depends on observation spacing, yet that spacing never appears in the input, in a positional embedding, or in a learned gate. Time is handled by the architecture's discretization, not by the network's representations.

\begin{figure}[h]

    \centering
    \definecolor{tidesGreen}{HTML}{3B6D11}
    \definecolor{tidesGreenFill}{HTML}{EAF3DE}
    \definecolor{tidesGreenText}{HTML}{173404}
    \definecolor{tidesGray}{HTML}{888780}
    \definecolor{tidesGrayFill}{HTML}{F1EFE8}
    \definecolor{idFill}{HTML}{EEEDFE}
    \definecolor{idText}{HTML}{26215C}
    \definecolor{s5Text}{HTML}{0C447C}
    \definecolor{mambaText}{HTML}{791F1F}
    
    \definecolor{physStroke}{HTML}{BA7517}
    \definecolor{physText}{HTML}{854F0B}
    
    \begin{tikzpicture}[
      font=\sffamily\footnotesize,
      >={Stealth[length=1.6mm,width=1.2mm]},
      every node/.append style={align=center},
      arr/.style       = {->, line width=0.4pt, black!75},
      box/.style       = {draw=tidesGray, line width=0.35pt,
                          fill=tidesGrayFill, rounded corners=2pt,
                          inner sep=2pt,
                          minimum width=22mm, minimum height=5.5mm},
      contrib/.style   = {box, fill=tidesGreenFill, draw=tidesGreen,
                          text=tidesGreenText},
      paneltitle/.style= {font=\footnotesize\itshape},
    ]
    
    \begin{scope}[local bounding box=panelS5]
      \node[paneltitle, font=\large\upshape\color{s5Text}] at (0, 0) {S5};
    
      \node[font=\footnotesize] (s-in) at (0,    -0.5) {input: $u$};
      \node[font=\footnotesize] (s-dt) at (1.5,  -0.5) {$\Delta$};
    
      \node[box, minimum height=11mm, text width=22mm]
            (s-pm) at (0, -1.4)
            {$\Lambda, B, C$ \\[-0.4ex] \scriptsize static};
    
      \node[box] (s-dc) at (0, -2.5) {Discretize};
      \node[box] (s-sc) at (0, -3.4) {Parallel scan};
      \node[box] (s-ot) at (0, -4.3) {Output};
    
      \draw[arr] (s-in.south) -- (s-pm.north);
      \draw[arr] (s-pm.south) -- (s-dc.north);
      \draw[arr, draw=physStroke, dashed] (s-dt.south) |- (s-dc.east);
      \draw[arr] (s-dc) -- (s-sc);
      \draw[arr] (s-sc) -- (s-ot);
    \end{scope}
    
    \begin{scope}[xshift=45mm, local bounding box=panelMamba]
      \node[paneltitle, font=\large\upshape\color{mambaText}] at (0, 0) {Mamba};
    
      \node[font=\footnotesize] (m-in) at (0,    -0.5) {input: $u$};
      \node[font=\footnotesize] (m-dt) at (1.5,  -0.5) {$\Delta$};

      \node[box] (m-cat) at (0, -1.1) {$h=\,$concat$(u,\Delta)$};
    
      \node[box, minimum height=11mm, text width=26mm, fill=idFill, text=idText]
            (m-pm) at (0, -2.2)
            {$\tilde\Delta(h), B(h),$ \\[-0.4ex] $C(h)$ \scriptsize projectors};
    
      \node[box] (m-dc) at (0, -3.3) {Discretize};
      \node[box] (m-sc) at (0, -4.2) {Parallel scan};
      \node[box] (m-ot) at (0, -5.1) {Output};
    
      \draw[arr] (m-in.south) -- (m-cat.north);
      \draw[arr, draw=physStroke, dashed] (m-dt.south) |- (m-cat.east);
      \draw[arr] (m-cat.south) -- (m-pm.north);
      \draw[arr] (m-pm.south) -- (m-dc.north);
      \draw[arr] (m-dc.south) -- (m-sc.north);
      \draw[arr] (m-sc) -- (m-ot);
    \end{scope}
    
    \begin{scope}[xshift=90mm, local bounding box=panelTIDES]
      \node[paneltitle, font=\large\upshape\color{tidesGreenText}, text=tidesGreenText] at (0, 0) {TIDES};
    
      \node[font=\footnotesize] (t-in) at (0,    -0.5) {input: $u$};
      \node[font=\footnotesize] (t-dt) at (1.5,  -0.5) {$\Delta$};
    
      \node[contrib, minimum height=11mm, text width=22mm, fill=idFill, text=idText]
            (t-pm) at (0, -1.4)
            {$\Lambda(u), B(u), C(u)$ \\[-0.4ex] \scriptsize projectors};
    
      \node[box] (t-dc) at (0, -2.5) {Discretize};
      \node[box] (t-sc) at (0, -3.4) {Parallel scan};
      \node[box] (t-ot) at (0, -4.3) {Output};
    
      \draw[arr] (t-in.south) -- (t-pm.north);
      \draw[arr] (t-pm.south) -- (t-dc.north);
      \draw[arr, draw=physStroke, dashed] (t-dt.south) |- (t-dc.east);
      \draw[arr] (t-dc) -- (t-sc);
      \draw[arr] (t-sc) -- (t-ot);
    \end{scope}
    
    \node[contrib, minimum width=3mm, minimum height=3mm, inner sep=0pt, 
    fill=idFill]
      at (-1.0, -5.8) (legmark) {};
    \node[anchor=west, font=\scriptsize, inner sep=1pt]
      at ([xshift=2mm]legmark.east)
      {\normalsize TIDES contribution (input-dependent components)};
    
\end{tikzpicture}

    \caption{Information flow through S5, Mamba, and TIDES architectures. $\Lambda$, $B$, and $C$ are static in S5, while Mamba makes $B$, $C$, and $\Tilde{\Delta}$ input-dependent. TIDES recovers implicit use of $\Delta$.}
    \label{fig:high_level_s5_mamba_tides}
\end{figure}
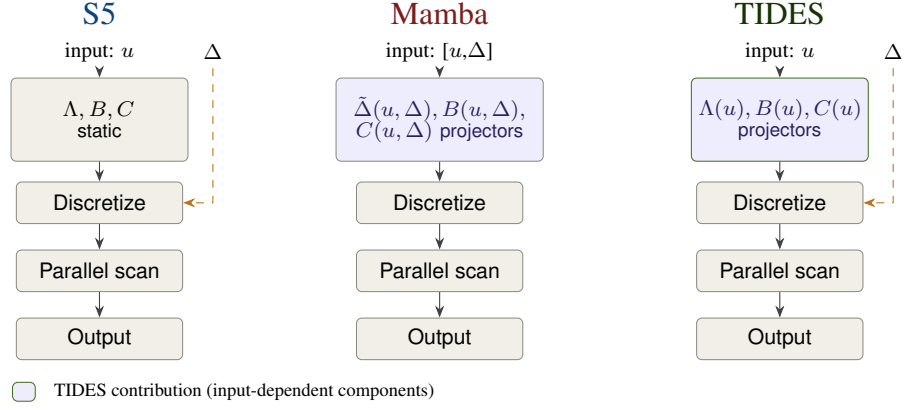

\paragraph{Contributions.}
\begin{itemize}
    \item \textbf{Architecture.} TIDES -- \textit{Time-Implicit Decay and Eigenvalue Selectivity}: a selective SSM variant with input-dependence on $(\mathrm{Re}(\lambda), B, C)$, a static $\mathrm{Im}(\lambda)$ and implicit $\Tilde{\Delta}$, designed so that physical sampling intervals enter the model only through the discretization (Section~\ref{sec:method}).
    \item \textbf{Mechanistic analysis.} A controlled toy experiment that isolates two orthogonal failure modes in existing SSMs -- S5's  linear time-invariant (LTI) rigidity and Mamba's failure to extrapolate across training $\Delta$ -- and demonstrates that TIDES is the only design overcoming both (Section~\ref{sec:toy}).
    \item \textbf{Empirical results.} New state-of-the-art, by average rank, on UEA time-series classification and Physiome-ODE regression benchmarks, and matching or exceeding the reference baseline model on 6 of 8 natively irregular datasets across astronomy, agriculture, neuromorphic sensing, and climate (Section~\ref{sec:experiments}).
\end{itemize}

\section{Background}
\label{sec:background}

\subsection{Continuous-time SSMs and discretization}

Linear SSMs are defined in continuous time by the equations below, 
where $H$ denotes the hidden (input/output) dimension and $P$ denotes the state dimension:
\begin{equation}
    \dot{x}(t) = A\,x(t) + B\,u(t), \qquad y(t) = C\,x(t) + D\,u(t),
\end{equation}
with input signal $u(t) \in \mathbb{R}^{H}$, latent state $x(t) \in \mathbb{C}^{P}$, and output $y(t) \in \mathbb{R}^{H}$. The model is parameterized by a state matrix $A \in \mathbb{C}^{P \times P}$ governing the latent dynamics, input and output matrices $B \in \mathbb{C}^{P \times H}$ and $C \in \mathbb{C}^{H \times P}$, and a feedthrough matrix $D \in \mathbb{R}^{H \times H}$.

To apply such a model to a discrete sequence $\{(t_k, u_k)\}_{k=1}^{L}$, the dynamics are integrated over each interval $\Delta_k := t_{k+1} - t_k$ to yield the recursion
\begin{equation}
    x_{k+1} = \bar{A}\,x_k + \bar{B}\,u_k, \qquad y_k = C\,x_k + D\,u_k,
\end{equation}
where the discrete matrices $(\bar{A}, \bar{B})$ are functions of $(A, B, \Delta)$ determined by the chosen discretization scheme such as bilinear, or zero-order hold (ZOH). A defining feature of this construction is that $\Delta$ enters the update as the \emph{physical} elapsed time between samples, i.e., doubling $\Delta$ produces the same state evolution as integrating the continuous system for twice as long. Discrete-time SSMs that preserve this property inherit a faithful notion of physical time from their continuous-time origin, a property we will return to in Section~\ref{sec:method} as the central design constraint motivating TIDES.

\subsection{S5: diagonal linear time-invariant dynamics}
S5~\citep{smith2023s5} parameterizes $A$ as a complex diagonal matrix $\Lambda = \mathrm{diag}(\lambda_1, \ldots, \lambda_P)$, where each $\lambda_p \in \mathbb{C}$ is a learned eigenvalue, initialized from a HiPPO-derived spectral decomposition~\citep{gu2020hippo}. Each eigenvalue defines a one-dimensional continuous-time mode: the real part $\mathrm{Re}(\lambda_p) < 0$ sets its decay rate (how quickly the mode forgets past inputs) and the imaginary part $\mathrm{Im}(\lambda_p)$ sets its oscillation frequency. Diagonality reduces the matrix exponential to elementwise scalars, so the recurrence admits a parallel associative scan for training efficiency~\citep{fisher1994parallelizing}. Since $\Lambda$ is fixed across the sequence, the resulting model is linear time-invariant (LTI): a given input contributes the same state increment regardless of context. Crucially for our setting, the integration step at step $k$, $\Delta_k$, enters S5 implicitly through the discretization, multiplying a learned but input-independent per-mode timescale, $\delta_p$; so irregular sampling is handled natively and the per-step variation of the effective step is set by the physical timestamps, not learned from the input.

\subsection{Mamba family: input-dependent selection}
Mamba \citep{gu2023mamba} and its variants Mamba-2 \citep{dao2024transformers} and Mamba-3 \citep{mamba3} (Appendix~\ref{app:mamba}) depart from the LTI regime by making the discretization parameters input-dependent. The recurrence is no longer time-invariant, as each token modulates its own dynamics, yet remains compatible with a parallel scan, which underlies Mamba's strong performance on long-range and language-modeling tasks. The cost is that $\tilde\Delta_k$ becomes a \emph{learned gate} rather than a physical sampling interval: Mamba assumes a uniform input grid, and actual elapsed time can only enter by concatenating $\Delta_k$ to $u_k$ as an input feature.

\section{Method}
\label{sec:method}

\subsection{Moving input-dependence from $\Delta$ to $\Lambda$}

Given the diagonal SSM recurrence under ZOH discretization
\begin{equation}
    x_{k+1} = \exp(\Lambda_k \, \Tilde{\Delta}_k) \, x_k + \bar{B}_k\, u_k, \qquad y_k = C_k x_k + D u_k,
\end{equation}
the design space of selectivity is the set of components that can be made input-dependent: $\Lambda$, $B$, $C$, and $\Tilde{\Delta}$ (Fig.~\ref{fig:high_level_s5_mamba_tides}). We argue for the following allocation:
\begin{itemize}
    \item \textbf{$\Tilde{\Delta}$ stays implicit and physical ($\Tilde{\Delta}\equiv\Delta$).} 
This gives the model its implicit time-awareness: the relationship between elapsed time and state evolution is fixed by the discretization rule \eqref{eq:zoh}, not learned from data.
    \item \textbf{$\mathrm{Re}(\Lambda)$ is input-dependent.} The decay rate of each mode becomes a function of the current input. Input-dependent decay has a clear interpretation: the model can ``forget faster'' when the input signals an event boundary, or ``hold longer'' when the input signals a quantity worth remembering. This is the closest semantic analog to Mamba's selectivity that does not conflict with the physical-time interpretation of $\Tilde{\Delta}$.
    \item \textbf{$\mathrm{Im}(\Lambda)$ stays static.} The imaginary part of each eigenvalue is the oscillation frequency of the corresponding mode. Per-token modulation of an oscillation frequency has no clean interpretation: the model would be redefining its own basis of dynamical modes at every step, and the resulting trajectory no longer corresponds to any coherent continuous-time signal. We empirically confirm in the ablation (Section~\ref{sec:ablation}) that input-dependent $\mathrm{Im}(\Lambda)$ hurts performance.
    \item \textbf{$B$ and $C$ are input-dependent.} These projections control how each token reads into the state and how the state is read out. Input-dependent $B,C$ provide additional expressivity in a way that is orthogonal to the eigenvalues of the dynamics; they allow filtering out noise and are analogous to a forget gate. We empirically confirm in the ablation (Section \ref{sec:ablation}) that input-dependence on $B$ and $C$ brings significant expressivity. 
\end{itemize}

\subsection{Architecture}

\paragraph{Selectivity heads.}
Three projection heads compute the input-dependent SSM parameters from each input $u_k \in \mathbb{R}^{H}$:
\begin{equation}
\label{eq:selectivity}
    \mathrm{Re}(\Lambda_k) = W_{\Lambda}\, u_k + \mathrm{Re}(\Lambda_0), \quad
    B_k = W_{B}\, u_k + B_0, \quad
    C_k = W_{C}\, u_k + C_0,
\end{equation}
where $\Lambda_0, B_0, C_0$ are the heads' bias terms, initialized to S5's HiPPO-derived values, and the projection weights $W_\Lambda, W_B, W_C$ are zero-initialized. With this initialization the model training begins from a well-conditioned HiPPO recurrence and learns selectivity as a smooth perturbation, rather than having to recover HiPPO from random initialization before the selection mechanism becomes useful.

\paragraph{Low-rank $B$ and $C$.}
A dense projection $W_B \in \mathbb{R}^{2PH \times H}$ has $2PH^2$ parameters; for typical settings ($P{=}128$, $H{=}128$), this single matrix exceeds the rest of the SSM in parameter count, leaving the model's capacity dominated by its selectivity heads on $B$ and $C$ rather than by the recurrence those heads modulate. We instead factor
\begin{equation}
    W_B = W_{B,\mathrm{up}}\, W_{B,\mathrm{down}}, \qquad W_{B,\mathrm{down}} \in \mathbb{R}^{r \times H},\;\; W_{B,\mathrm{up}} \in \mathbb{R}^{2PH \times r},
\end{equation}
and analogously for $C$, with rank $r$ a hyperparameter. This low-rank structure also provides regularization by bottle-necking these large projectors, which can be adjusted with the $r$ knob. Only $W_{B,\mathrm{up}}$ and $W_{C,\mathrm{up}}$ are zero-initialized, which preserves the HiPPO biases at step zero. The per-head cost drops from $\mathcal{O}(PH^2)$ to $\mathcal{O}(rPH)$. The $\Lambda$ head is left full-rank, as its output dimension is $P$ rather than $PH$.  

\paragraph{Deep projectors.}
The heads in \eqref{eq:selectivity} are affine in $u_k$. Mamba enriches its
input pathway with a depthwise convolution, a SiLU nonlinearity, and
multiplicative gating before computing its selection parameters; we add
comparable nonlinear preprocessing by composing $d$ residual gated-linear-unit (GLU) blocks before the final projection. Concretely, define
\begin{equation}
    g^{(d)}(x) = W_{\mathrm{out}}\bigl(b_d \circ \cdots \circ b_1\bigr)(x),
    \qquad
    b_k(x) = x + (W^{(1)}_k x) \odot \sigma(W^{(2)}_k x),
\end{equation}
where $W_{\mathrm{out}}$ projects from width $H$ to the target dimension.
The $\Lambda$ head replaces $u_k$ with $g^{(d_\Lambda)}(u_k)$ before the affine
map, and the input encoder applies $g^{(d_{\mathrm{enc}})}(\cdot)$ with
$W_{\mathrm{out}} : \mathbb{R}^{d_{\mathrm{input}}} \to \mathbb{R}^{H}$.
Setting $d = 0$ reduces $g^{(d)}$ to a plain linear map, recovering the affine
baseline; $d_\Lambda = d_{\mathrm{enc}} = 0$ is our default. In practice, we set $d_\Lambda$ and $d_{\mathrm{enc}}$ as hyperparameters.

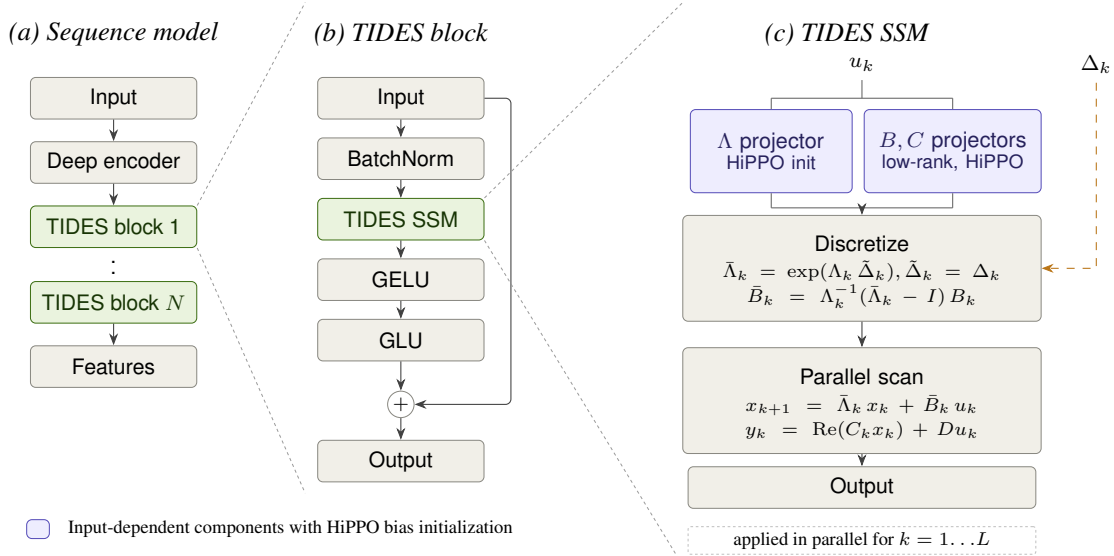
\begin{figure}[t]
    \definecolor{tidesGreen}{HTML}{3B6D11}
    \definecolor{tidesGreenFill}{HTML}{EAF3DE}
    \definecolor{tidesGreenText}{HTML}{173404}
    \definecolor{tidesGray}{HTML}{888780}
    \definecolor{tidesGrayFill}{HTML}{F1EFE8}
    \definecolor{idFill}{HTML}{EEEDFE}
    \definecolor{idStroke}{HTML}{534AB7}
    \definecolor{idText}{HTML}{26215C}
    \definecolor{physStroke}{HTML}{BA7517}
    \centering
    \begin{tikzpicture}[
      font=\sffamily\footnotesize,
      >={Stealth[length=1.6mm,width=1.2mm]},
      every node/.append style={align=center},
      arr/.style       = {->, line width=0.4pt, black!75},
      flow/.style      = {line width=0.4pt, black!60},
      zoom/.style      = {dashed, dash pattern=on 1.2pt off 1.2pt,
                          line width=0.35pt, black!40},
      box/.style       = {draw=tidesGray, line width=0.35pt,
                          fill=tidesGrayFill, rounded corners=2pt,
                          inner sep=2pt,
                          minimum width=22mm, minimum height=5.5mm},
      contrib/.style   = {box, fill=tidesGreenFill, draw=tidesGreen,
                          text=tidesGreenText},
      paneltitle/.style= {font=\normalsize\itshape},
    ]

    \begin{scope}[local bounding box=panela]
      \node[paneltitle] at (0,0) {(a) Sequence model};
      \node[box] (a-input) at (0,-0.85) {Input};
      \node[box] (a-enc)   at (0,-1.70) {Deep encoder};
      \node[contrib] (a-b1)    at (0,-2.55) {TIDES block 1};
      \node                    (a-dots)  at (0,-3.05) {$\vdots$};
      \node[contrib] (a-bN)    at (0,-3.55) {TIDES block $N$};
      \node[box] (a-feat)  at (0,-4.40) {Features};

      \draw[arr] (a-input) -- (a-enc);
      \draw[arr] (a-enc)   -- (a-b1);
      \draw[arr] (a-bN)    -- (a-feat);
    \end{scope}

    \begin{scope}[xshift=38mm, local bounding box=panelb]
      \node[paneltitle] at (0,0) {(b) TIDES block};
      \node[box] (b-in)   at (0,-0.85) {Input};
      \node[box] (b-bn)   at (0,-1.65) {BatchNorm};
      \node[contrib] (b-ssm)  at (0,-2.45) {TIDES SSM};
      \node[box] (b-gelu) at (0,-3.25) {GELU};
      \node[box] (b-glu)  at (0,-4.05) {GLU};
      \node[circle, draw=tidesGray, line width=0.35pt, inner sep=0pt,
            minimum size=3.5mm, font=\scriptsize] (b-add) at (0,-4.90) {$+$};
      \node[box] (b-out)  at (0,-5.65) {Output};

      \draw[arr] (b-in)   -- (b-bn);
      \draw[arr] (b-bn)   -- (b-ssm);
      \draw[arr] (b-ssm)  -- (b-gelu);
      \draw[arr] (b-gelu) -- (b-glu);
      \draw[arr] (b-glu)  -- (b-add);
      \draw[arr] (b-add)  -- (b-out);

      \draw[arr, rounded corners=2pt]
        (b-in.east) -- ++(3.5mm,0) |- (b-add.east);
    \end{scope}

    \draw[zoom] (a-b1.north east) -- ([yshift=1.5mm]panelb.north west);
    \draw[zoom] (a-b1.south east) -- ([yshift=-1.5mm]panelb.south west);

    \begin{scope}[xshift=82mm, local bounding box=panelc]
      \node[paneltitle] at (1.5, 0) {(c) TIDES SSM};

      \node[font=\footnotesize] (c-ut) at (1.7, -0.4) {$u_k$};
      \node[font=\footnotesize] (c-dt) at (4.8, -0.4) {$\Delta_k$};

      \node[box, draw=idStroke, fill=idFill, text=idText,
            minimum width=20mm, minimum height=11mm, text width=20mm]
            (c-lam) at (0.5, -1.55)
            {$\Lambda$ projector \\[-0.4ex] \scriptsize HiPPO init};
      \node[box, draw=idStroke, fill=idFill, text=idText,
            minimum width=22mm, minimum height=11mm, text width=22mm]
            (c-bc)  at (2.9, -1.55)
            {$B,C$ projectors \\[-0.4ex] \scriptsize low-rank, HiPPO};

      \draw[flow] (c-ut.south) -- (c-ut |- 0,-0.85);
      \draw[flow] (c-lam.north |- 0,-0.85) -- (c-bc.north |- 0,-0.85);
      \draw[flow] (c-lam.north |- 0,-0.85) -- (c-lam.north);
      \draw[flow] (c-bc.north  |- 0,-0.85) -- (c-bc.north);

      \draw[flow] (c-lam.south) -- (c-lam.south |- 0,-2.30);
      \draw[flow] (c-bc.south)  -- (c-bc.south  |- 0,-2.30);
      \draw[flow] (c-lam.south |- 0,-2.30) -- (c-bc.south |- 0,-2.30);

      \node[box, minimum width=46mm, minimum height=14mm, text width=46mm]
            (c-disc) at (1.7, -3.10)
            {\rule{0pt}{2ex}Discretize \\[0.5ex] \scriptsize
            $\bar{\Lambda}_k = \exp(\Lambda_k\, \Tilde{\Delta}_k), \Tilde{\Delta}_k=\Delta_k$ \\[-0.1ex]
            $\bar{B}_k = \Lambda_k^{-1}(\bar{\Lambda}_k - I)\, B_k$};

      \node[box, minimum width=46mm, minimum height=14mm, text width=46mm]
            (c-scan) at (1.7, -4.85)
            {\rule{0pt}{2ex}Parallel scan \\[0.5ex] \scriptsize
            $x_{k+1} = \bar{\Lambda}_k\, x_k + \bar{B}_k\, u_k$ \\[-0.1ex]
            $y_k = \mathrm{Re}(C_k x_k) + D  u_k$};

      \node[box, minimum width=46mm] (c-out) at (1.7, -6.00) {Output};

      \draw[arr] (1.7, -2.30) -- (c-disc.north);

      \draw[arr, draw=physStroke, dashed] (c-dt.south) |- (c-disc.east);

      \draw[arr] (c-disc) -- (c-scan);
      \draw[arr] (c-scan) -- (c-out);

      \node[draw=black!25, dashed, dash pattern=on 1pt off 1pt,
            line width=0.3pt, rounded corners=1pt, inner sep=2pt,
            minimum width=46mm, font=\scriptsize]
            at (1.7,-6.70)
            {applied in parallel for $k = 1{\ldots}L$};
    \end{scope}

    \draw[zoom] (b-ssm.north east) -- ([yshift=1.5mm]panelc.north west);
    \draw[zoom] (b-ssm.south east) -- ([yshift=-1.5mm]panelc.south west);

    \node[box, fill=idFill, draw=idStroke,
          minimum width=3mm, minimum height=3mm, inner sep=0pt]
      at (-1.0,-6.55) (legmark) {};
    \node[anchor=west, font=\scriptsize, inner sep=1pt]
      at ([xshift=2mm]legmark.east)
      {\small Input-dependent components with HiPPO bias initialization};

    \end{tikzpicture}
    \caption{TIDES architecture: from sequence model (a) to TIDES block (b) to the lower-level SSM (c). In (c), only the ZOH discretization is shown explicitly, but both ZOH and bilinear are possible. }
    \label{fig:tides_architecture}
\end{figure}

\paragraph{Projector normalization.}
In the LTI case, $\Lambda$, $B$, and $C$ are static parameters. When these quantities become 
input-dependent, unconstrained projectors produce them at every 
timestep, introducing perturbations that are harder to bound: errors in 
$\mathrm{Re}(\Lambda_k)$ are exponentially amplified through the recurrence, 
while $B_k$ affects the state norm at every step. To stabilize 
training under this harder regime, we apply RMSNorm to the projected 
$\mathrm{Re}(\Lambda_k)$ and a complex-valued variant to the projected $B_k$ 
and $C_k$ immediately before discretization. This is analogous to the BCNorm 
adopted in Mamba-3~\citep{mamba3}, which applies RMSNorm to the 
projected $B$ and $C$ activations for the same reason. Empirically, the 
normalization matters most when deep projectors ($d_\Lambda > 0$) are used, 
where the projection itself can accumulate scale (Fig.~\ref{fig:tides_architecture}).

\paragraph{$\Lambda$ reparameterization.}
SSMs are known to benefit from reparameterizing $\mathrm{Re}(\Lambda)$ rather
than learning it directly. Common choices include the exponential
parameterization $\mathrm{Re}(\Lambda) = -\exp(\theta)$, and the
stable parameterization $\mathrm{Re}(\Lambda) = -1/(\theta^2 + 1/2)$
\cite{stablessm}. These formulas map an unconstrained $\theta \in \mathbb{R}$
to a strictly negative decay rate, which keeps $\Lambda$ in the stable
half-plane throughout training and prevents gradient descent from pushing
modes onto the stability boundary, where gradients across decay rates become
poorly conditioned and the model is effectively confined to exponentially
decaying memory. The same reasoning applies, and arguably more strongly, when
$\mathrm{Re}(\Lambda)$ is input-dependent: a stable parameterization ensures
that no token can produce a locally explosive recurrence, regardless of where
its projection lands. In practice, we treat the reparameterization as a hyperparameter.

\section{The \textit{Fading Flash} experiment: two failure modes}
\label{sec:toy}

\begin{wrapfigure}{l}{0.53\linewidth}           
      \vspace{-1.0em}                         
      \centering                       
      \includegraphics[width=0.85\linewidth]{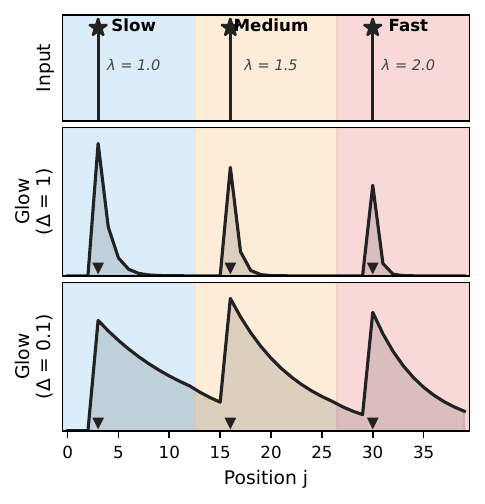}             
      \vspace{-0.5em}                 
      \caption{\textbf{Task setup.} 40 detectors split into three rate zones. Sparse flashes (top) produce zone-dependent decaying glows; the same flashes under different $\Delta$ (middle, bottom) yield rescaled dynamics. The model needs to predict the correct decaying glows given the sparse flashes, zone and $\Delta$ values.}
      \label{fig:toy_setup}            
      \vspace{-4.0em}
\end{wrapfigure}

Before scaling to large benchmarks, we use a controlled toy problem to better illustrate our design argument (Fig.~\ref{fig:toy_setup}). The setup: 
a row of 40 detectors is hit by sparse flashes; each detector glows and exponentially fades after a hit; the detectors are partitioned into three colored zones with different fade rates (slow, medium, fast). A still-glowing detector that crosses into a new zone takes on that zone's fade rate. A global ``clock time'' $\Delta_{k=1, ..., 40}=\Delta$ stretches or compresses the entire trajectory.

This setup isolates the two architectural axes:
\begin{itemize}
    \item \emph{LTI vs Input-dependent:} A purely LTI model has a single fixed readout and cannot represent three different decay rates simultaneously.
    \item \emph{How is $\Delta$ exposed to the model?} A model that learns $\Delta$ as a function of the input cannot extrapolate to clock speeds outside the training range.
\end{itemize}

We train three SSM core variants: S5 as LTI baseline, Mamba$_{\text{S}}$ (a Mamba surrogate where we take the vanilla SSM core and make $\Tilde{\Delta}, B, C$ input-dependent), TIDES (input-dependent $\mathrm{Re}(\Lambda), B, C$). For Mamba, we input $\Delta$ as a separate input channel, following the standard practice when irregular timestamps cannot be implicitly baked into the model. All models are matched in parameter count.

\begin{figure}[h]
    \centering
    \includegraphics[width=\linewidth]{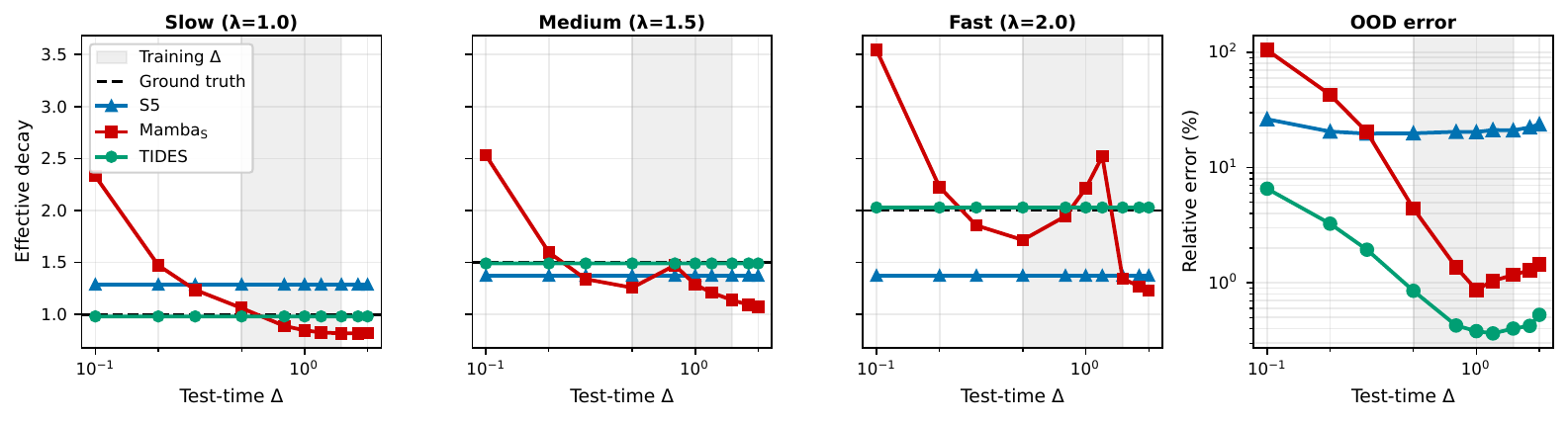}
    \caption{Left: \textbf{Effective learned decay vs.\ test $\Delta$, per zone.} S5 and TIDES bake $\Delta$ into the discretization, so the learned decay stays flat across the full test range. Mamba$_{\text{S}}$ distorts the physically meaningful $\Delta$ through a learned gate, breaking the cancellation and causing drift outside training. Right: \textbf{Relative error vs.\ test $\Delta$.} TIDES consistently outperforms Mamba$_{\text{S}}$ and S5 both in- and out-of-distribution. See Appendix \ref{app:toy} for full details.}
    \label{fig:toy_mechanism}
\end{figure}

\paragraph{Findings.} The three models split cleanly along two axes: \emph{Expressivity} (whether the model represents zone-conditional decay) and \emph{Extrapolation} (whether its behavior extrapolates to unseen $\Delta$).

\emph{S5 lacks expressivity.} S5's effective learned decays stay constant per zone, and its relative error stays robust across unseen $\Delta$ values (Fig.~\ref{fig:toy_mechanism}), thanks to its implicit use of $\Delta$. However, without input-dependence, no single linear readout can reproduce three different zone decays simultaneously, and the model plateaus at the LTI floor, failing to capture the zone-dependent decays.

\emph{Mamba lacks extrapolation.} The learned gate $\tilde\Delta_k = \mathrm{softplus}(W_\Delta[u_k, \Delta])$ fits the training $\Delta$ distribution well, achieving low training error; however, this mechanism precludes implicit time-awareness. As a result, effective decay rates drift with $\Delta$ and error grows sharply outside the training range (Fig.~\ref{fig:toy_mechanism}).

\emph{TIDES overcomes both.} Input-dependence on $\Lambda$ provides expressivity, while baking $\Delta$ in implicitly provides extrapolation across unseen timedeltas by construction. Eigenvalues land at the three target decays, and the model stays robust across a wide span of $\Delta$.

\section{Experiments}
\label{sec:experiments}

We compare TIDES to S5 \citep{smith2023s5}, Mamba/S6 \citep{gu2023mamba}, Mamba-2 \citep{dao2024transformers} and Mamba-3 \citep{mamba3}, Rough Transformer \citep{moreno2024rough}, and the rest of the irregular-time-series baselines: LRU \citep{orvieto2023resurrecting}, NCDE \citep{kidger2020neural}, NRDE \citep{morrill2021neural}, LogNCDE \citep{walker2024log}, and a vanilla Transformer \citep{vaswani2017attention} on UEA, and GRU-ODE-Bayes \citep{debrouwer2019gruode}, Neural Flows \citep{bilos_neural_2021}, CRU \citep{cru_schirmer2022modeling}, LinODEnet \citep{scholz2023latent}, and GraFITi/GraFITi-C \citep{yalavarthi2024grafiti} on Physiome-ODE.
TIDES is implemented in PyTorch. Hyperparameters are tuned per-dataset within a fixed budget. Full search spaces, the justification behind the dataset choices, and the complete results are given in Appendix \ref{app:experiments}. All experiments use approximately 5,000 GPU-hours of A100 time.

\subsection{UEA time series classification}
\label{sec:uea_timeseries}
For time series classification, we select the UEA multivariate archive \cite{bagnall2018uea} and follow the Rough Transformer environment \citep{moreno2024rough}. Complete dataset and experiment details are given in Appendix \ref{app:uea}. TIDES achieves the highest average accuracy across the six UEA datasets, surpassing Mamba-3 by approximately 1.5 points on average and ranking first on 2 of 6 datasets (Table \ref{tab:uea}), demonstrating its capability to also model sequences with uniform gaps.

\vspace{-4mm}
\begin{table}[h!]
\centering
\caption{UEA time series classification results (accuracy \% $\pm$ standard deviation) for select baselines (truncated from Appendix~\ref{app:uea_results}). Bold indicates the best result per dataset. Baseline accuracies are reproduced from Walker et al. \cite{walker2024log} and Moreno et al. \cite{moreno2024rough}.}
\label{tab:uea}
\footnotesize
\resizebox{\linewidth}{!}{%
\begin{tabular}{lcccccccc}
\toprule
Dataset & TIDES & Mamba-3 & LogNCDE & RFormer & S6 & LRU & S5 & Transformer\\
\midrule
SCP1 & $87.6{\pm}3.3$ & $83.3{\pm}2.8$ & $83.1{\pm}2.8$ & $81.2{\pm}2.8$ & $82.8{\pm}2.7$ & $82.6{\pm}3.4$ & $\mathbf{89.9{\pm}4.6}$ & $84.3{\pm}6.3$ \\
SCP2 & $\mathbf{56.5{\pm}6.2}$ & $55.4{\pm}3.6$ & $53.7{\pm}4.1$ & $52.3{\pm}3.7$ & $49.9{\pm}9.5$ & $51.2{\pm}3.6$ & $50.5{\pm}2.6$ & $49.1{\pm}2.5$ \\
MI   & $\mathbf{58.2{\pm}2.8}$ & $57.9{\pm}5.4$ & $53.7{\pm}5.3$ & $55.8{\pm}6.6$ & $51.3{\pm}4.7$ & $48.4{\pm}5.0$ & $47.7{\pm}5.5$ & $50.5{\pm}3.0$ \\
EW   & $85.7{\pm}4.0$ & $84.0{\pm}2.6$ & $85.6{\pm}5.1$ & $\mathbf{90.3{\pm}0.1}$ & $85.0{\pm}16.1$ & $87.8{\pm}2.8$ & $81.1{\pm}3.7$ & OOM \\
ETC  & $34.2{\pm}2.4$ & $34.9{\pm}4.8$ & $34.4{\pm}6.4$ & $34.7{\pm}4.1$ & $26.4{\pm}6.4$ & $21.5{\pm}2.1$ & $24.1{\pm}4.3$ & $\mathbf{40.5{\pm}6.3}$ \\
HB   & $74.4{\pm}4.0$ & $72.8{\pm}4.3$ & $75.2{\pm}4.6$ & $72.5{\pm}0.1$ & $76.5{\pm}8.3$ & $\mathbf{78.4{\pm}6.7}$ & $77.7{\pm}5.5$ & $70.5{\pm}0.1$ \\
\midrule
Avg      & $\mathbf{66.1}$ & 64.7 & 64.3 & 64.5 & 62.0 & 61.7 & 61.8 & 59.0 \\
Avg Rank & $\mathbf{3.00}$ & 4.17 & 4.25 & 5.17 & 6.33 & 6.50 & 7.08 & 7.50 \\
\bottomrule
\end{tabular}%
}
\end{table}

\subsection{Physiome-ODE}
Physiome-ODE is a comprehensive, state-of-the-art regression benchmark for IMTS, comprised of 50 irregular-time-series forecasting datasets derived from biophysical models. We follow the protocol of Klötergens et al. \citep{klotergens2025physiome} -- the complete details are given in Appendix \ref{app:physiome}. TIDES attains the best average rank and is statistically on par with the strongest continuous-time baseline, LinODEnet, tying it on per-dataset wins (16/50) (Table \ref{tab:physiome_ode_results}. A bootstrap 95\% CI on the average-rank difference between TIDES and LinODEnet is [-0.87, +0.34], which includes zero, albeit skewed in TIDES’s favor. The full per-dataset results are given in Appendix \ref{app:physiome}). 

\begin{table}[h!]
\centering
\caption{Physiome-ODE benchmark results. Full per-dataset results are given in Appendix \ref{app:physiome}. Baseline MSEs are reproduced from \citet{klotergens2025physiome}. TIDES is our run.}
\footnotesize
\resizebox{\linewidth}{!}{%
\begin{tabular}{lccccccc}
\toprule
Dataset & TIDES & LinODEnet & GraFITi & GraFITi-C & CRU & Neural Flows & GRU-ODE-Bayes \\
\midrule
\# Wins & $\mathbf{16}$ & $\mathbf{16}$ & 5 & 10 & 3 & 0 & 0 \\
Avg.\ Rank & $\mathbf{2.40}$ & 2.62 & 2.82 & 3.54 & 3.68 & 5.76 & 6.88 \\
\bottomrule
\end{tabular}%
}
\label{tab:physiome_ode_results}
\end{table}

\subsection{Random drop}
\label{sec:ablation}

\vspace{-3mm}
To isolate the effect of the sampling regime and assess generalization to irregular sampling in a controlled way, we vary observation density on a real dataset while holding everything else fixed. This \emph{random drop} experiment is tested on EigenWorms: a different random 50\% of timesteps is dropped per seed; evaluation uses fixed indices at varying $r_\mathrm{test}$, with original timestamps preserved to expose the elapsed gap.  We train six $\sim\!30$k-parameter SSM variants that differ only in which of $\{\mathrm{Re}(\Lambda), \mathrm{Im}(\Lambda), B{,}C, \Tilde{\Delta} \}$ is input-dependent (Appendix \ref{app:ablations}, Table \ref{tab:droprate_full}), in addition to three official Mamba variants. RFormer \citep{moreno2024rough} is included as a strong non-SSM continuous-time baseline.

   \begin{wrapfigure}{r}{0.55\linewidth}
    \vspace{-1.0em}
    \centering                           
    \includegraphics[width=\linewidth]{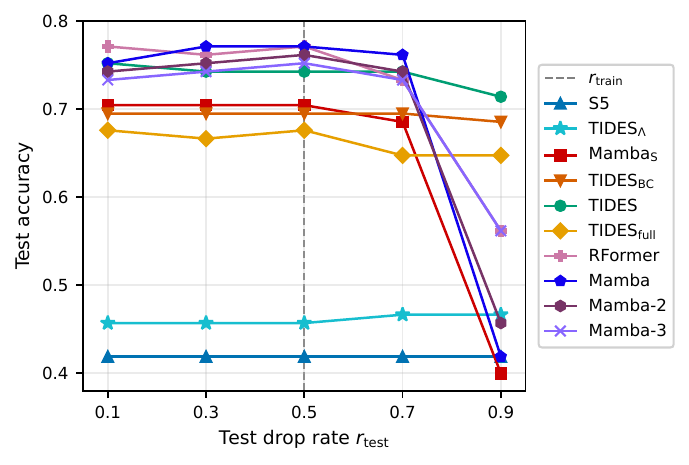}        
    \vspace{-1.5em}                                   
    \caption{Test accuracy vs.\ $r_\mathrm{test}$ on EigenWorms ($n{=}3$ seeds; trained at $r_\mathrm{train}{=}0.5$).}                                  
    \label{fig:droprate}
    \vspace{-1.0em}                
\end{wrapfigure}

In Fig.~\ref{fig:droprate}, we show that (i) LTI $B,C$ caps expressivity, as S5 and TIDES$_\Lambda$ plateau near chance regardless of $r_\mathrm{test}$. (ii) Input-dependent $\tilde\Delta$ does not extrapolate: Mamba, Mamba-2, and Mamba-3 match TIDES for $r_\mathrm{test}\le0.7$ but collapse from $\sim$0.7 to $\sim$0.4-0.6 at $r_\mathrm{test}{=}0.9$; RFormer fails analogously, with its fixed signature window count baked into positional encodings. (iii) Input-dependent $\mathrm{Re}(\Lambda)$ recovers expressivity \emph{and} extrapolation: TIDES achieves the best mean accuracy ($0.739$) and stays flat across the full $r_\mathrm{test}$ range. Adding input-dependence to $\mathrm{Im}(\Lambda)$ (TIDES$_\mathrm{full}$) does not help and slightly hurts, confirming that selectivity belongs on decay rates, not on oscillation frequencies.

\subsection{Natively irregular benchmarks}

\label{sec:irregular}
To test TIDES where irregular sampling is a property of the measurement process itself, we evaluate on real-world data from four domains, where the observation times are set by the instrument or the environment: ground-based telescope light curves of periodic variable stars from MACHO~\citep{naul2018recurrent} and ASAS-SN~\citep{jayasinghe2018asassn,jayasinghe2019asassn}, whose cadence follows observing conditions; Sentinel-2 satellite crop time series from TimeSen2Crop~\citep{weikmann2021timesen2crop} and EuroCropsML~\citep{reuss2025eurocropsml}, where revisit times and cloud cover give every sample its own set of dates; neuromorphic sensor recordings from SHD and SSC~\citep{cramer2022heidelberg} and DVS128 Gesture~\citep{amir2017low}, whose asynchronous events arrive at intervals spanning orders of magnitude; and weather station records from USHCN~\citep{menne_ushcn}, where each variable is reported at its own times. Each dataset follows the evaluation protocol of the paper that reports its baseline (Table~\ref{tab:additional_results}).

\begin{table}[h]
  \centering
  \caption{TIDES against the reference baseline on natively irregular benchmarks. Each dataset follows the evaluation protocol of its source paper. Standard deviations are shown where both TIDES and the baseline report them. Arrows give the direction of improvement; best result per row in bold. EuroCropsML uses the 500-shot Estonia setting of DirPA.}
  \label{tab:additional_results}
  \small
  \setlength{\tabcolsep}{5pt}
  \begin{tabular}{@{}lllccl@{}}
    \toprule
    \textbf{Domain} & \textbf{Dataset} & \textbf{Metric} & \textbf{TIDES} & \textbf{Baseline} & \textbf{Baseline model} \\
    \midrule
    \multirow{2}{*}{Astronomy}
      & MACHO        & Acc.\ (\%) $\uparrow$ & $\mathbf{92.8}$                               & $92.7$                                & iTCN~\citep{itcn} \\
      & ASAS-SN      & Acc.\ (\%) $\uparrow$ & $94.2$                                        & $\mathbf{94.5}$                       & iTCN~\citep{itcn} \\
    \addlinespace
    \multirow{2}{*}{Agriculture}
      & TimeSen2Crop & OA (\%) $\uparrow$    & $\mathbf{89.80}$                               & $88.70$                               & MLP+LTAE~\citep{vincent2023pixelwise} \\
      & EuroCropsML  & Acc.\ (\%) $\uparrow$ & $\mathbf{80.26}_{\scriptscriptstyle\pm 0.41}$  & $79.14_{\scriptscriptstyle\pm 0.23}$  & DirPA~\citep{dirpa} \\
    \addlinespace
    \multirow{3}{*}{Neuromorphic}
      & SHD          & Acc.\ (\%) $\uparrow$ & $\mathbf{96.1}$                               & $95.9$                                & Event-SSM~\citep{eventssm} \\
      & SSC          & Acc.\ (\%) $\uparrow$ & $\mathbf{88.6}$                               & $88.4$                                & Event-SSM~\citep{eventssm} \\
      & DVS128       & Acc.\ (\%) $\uparrow$ & $98.9$                                        & $\mathbf{99.2}^{\dagger}$             & Event-SSM~\citep{eventssm} \\
    \addlinespace
    \multirow{2}{*}{Climate}
      & USHCN        & MSE $\downarrow$      & $\mathbf{0.147}_{\scriptscriptstyle\pm 0.003}$ & $0.159_{\scriptscriptstyle\pm 0.014}$ & APN~\citep{apn} \\
      & USHCN        & MAE $\downarrow$      & $\mathbf{0.211}_{\scriptscriptstyle\pm 0.002}$ & $0.261_{\scriptscriptstyle\pm 0.017}$ & APN~\citep{apn} \\
    \bottomrule
  \end{tabular}

  \vspace{3pt}
  \begin{minipage}{0.92\linewidth}
    \footnotesize $^{\dagger}$Updated result from the authors' code release (v0.2) after a fix to the DVS event tokenization~\citep{eventssm_repo}; the paper reports 97.7.
  \end{minipage}
\end{table}

USHCN is asynchronous across channels: each variable is observed at its own times. We therefore convert it from wide format, where each timestamp holds a value or a missing entry for every channel, to long format, where each observation is a triplet $(t, c, x)$ of timestamp, channel index and value, respectively, so that only observed entries are kept. 
TIDES sets the best reported result on TimeSen2Crop, EuroCropsML ($p{=}0.0015$) and USHCN, and exceeds Event-SSM, whose pipeline it plugs into, on SHD and SSC. On the variable-star benchmarks, it matches iTCN on MACHO and is slightly below it on ASAS-SN, with mean paired differences over the 8 splits of $+0.13{\pm}0.32$ and $-0.36{\pm}0.17$ (computed on unrounded per-split accuracies), while outperforming every other model reported by \citet{itcn}. On DVS128, it is slightly below the updated Event-SSM result.

\section{Related work}
\label{sec:related}

\paragraph{State space models.} S4 \citep{gu2022s4} introduced structured SSMs. S5 \citep{smith2023s5} simplified to complex-diagonal form with a per-step $\Tilde{\Delta}_k$ amenable to parallel scan. Mamba/S6 \citep{gu2023mamba} introduced selectivity over $(\Tilde{\Delta}, B, C)$. A parallel line of gated linear recurrences --- GLA \citep{yang2023gated}, Mamba-2 \citep{dao2024transformers}, Mamba-3 \citep{mamba3}  --- also injects input-dependence into the state transition via data-dependent scalar or structured decay, but operates purely discretely: there is no physically-meaningful $\Delta$, so irregular timestamps must enter as an input feature. TIDES is, to the best of our knowledge, the first to combine selective $\Lambda$ with a physical-time discretization.

\paragraph{Irregular time series.}
Methods for IMTS handle elapsed time explicitly, through one of three mechanisms. \emph{Numerical integration between observations:} Latent ODE \citep{rubanova2019latent} and ODE-RNN propagate the hidden state with a neural ODE solver, while CRU \citep{cru_schirmer2022modeling} replaces the ODE with a linear SDE that admits closed-form Kalman-style updates between observations. \emph{Time as a gated input:} GRU-D \citep{grud_che2018recurrent} feeds elapsed time into a learned exponential decay on the hidden state. \emph{Time as a positional code:} mTAN \citep{mtan_shukla2021multi} parameterizes attention with a kernel over elapsed time, and ContiFormer \citep{chen2023contiformer} and Rough Transformer \citep{moreno2024rough} extend transformers with continuous-time machinery (ODE-driven attention and path signatures, respectively).

\paragraph{Continuous-time deep learning.} Neural ODEs \citep{chen2018neuralode} and neural CDEs \citep{kidger2020neural} formulate sequence models as ODE solvers. These methods provide flexible time handling but incur a substantial computational cost compared to SSM-style discretization.

\section{Discussion}
\label{sec:discussion}

\paragraph{Interpreting the empirical pattern.}
The advantage TIDES holds over each baseline scales with how much the task benefits from three properties: per-token input-dependent processing needed to filter noise and resolve context, faithful handling of irregular sampling intervals, and extrapolation to sampling regimes not seen during training. Different baselines satisfy different subsets, which explains the variation in the margin we observe. On UEA, where the dominant competitor Mamba-3 has no implicit continuous-time discretization, TIDES wins on average rank and accuracy. On Physiome-ODE, the strongest baseline LinODEnet is itself continuous-time, sharing two of the three properties; the gap narrows accordingly, with TIDES retaining a small edge on average rank ($2.40$ vs.\ $2.62$) and tying on per-dataset wins (16/50). The third axis, namely extrapolation across irregularity is what the \emph{Fading Flash} diagnostic and the random-drop ablation (Section~\ref{sec:ablation}) isolate in controlled settings: TIDES is the only architecture in our comparison that satisfies all three properties simultaneously, while ablations such as Mamba$_{\text{S}}$ and RFormer collapse precisely when the test-time sampling regime diverges from training.

\paragraph{A recipe for expressive irregular sequence modeling.}
Beyond TIDES, our analysis surfaces a design principle that broadly applies to architectures for irregularly-sampled sequences: keep discretization and expressivity separate. The physical sampling interval $\Delta_k$ should enter the model only through the rule that converts the continuous-time generator into a discrete recurrence, where its semantics are fixed analytically rather than learned. Expressivity should live on the continuous-time generator itself, in our case $(\mathrm{Re}(\Lambda), B, C)$. This separation is what allows a model to be simultaneously expressive (through its input-dependent components) and extrapolative across sampling regimes (through the analytic time-handling of the discretization). When the two roles are conflated, as in the case of Mamba, where the discretization step is itself made a learned function of the input, the model recovers per-token expressivity but loses extrapolation, as the \emph{Fading Flash} experiment illustrates. 

\paragraph{Scope.}
TIDES is mainly designed for sequence problems where the spacing between observations carries information such as irregular clinical, simulation, or sensor-based time series. We do not claim it as a general-purpose replacement for Mamba on regular-grid tasks such as language modeling, where the input-dependent $\tilde\Delta$ gate plays a different role and the implicit-time benefit does not apply. Whether the principle survives at language-model scale is an open question we leave to future work.

\paragraph{Limitations.}
We do not evaluate TIDES on language modeling or other regular-grid long-range benchmarks, so our results speak mainly to the irregular-time regime. The \emph{Fading Flash} experiment is a controlled diagnostic designed to isolate failure modes; it is not itself evidence of real-world advantage, which we instead derive from UEA and Physiome-ODE. TIDES introduces additional tunable structure beyond S5 such as rank $r$ for the low-rank $B$ and $C$ projectors, projector depths $d_\Lambda$ and $d_\mathrm{enc}$, and the choice of $\Lambda$ reparameterization, potentially posing more tuning work. Our implementation uses a generic parallel scan rather than a hardware-aware kernel, so the wall-clock comparisons should be read as architectural rather than fully optimized. Our argument that input-dependent $\mathrm{Im}(\Lambda)$ is a poor selectivity target is conceptual and limited in the empirical validation (Section~\ref{sec:ablation}), we do not provide a formal expressivity result.

\paragraph{Broader impact.}
TIDES inherits the SSM family's linear training scaling and constant per-step inference cost in recurrent mode, and adds native handling of irregular timestamps without explicit time features. The combination is well-suited to real-time sensor settings on resource-constrained hardware such as wearable health monitoring, where streams arrive at irregular intervals, sensors drop and resume, and compute and memory budgets are tight. The implicit time-awareness we add lets such systems operate without engineering an explicit time-encoding pipeline or maintaining separate logic for missing observations, and the extrapolation behavior we demonstrate offers some robustness to sampling regimes that differ from those seen at training time.

\paragraph{Future directions.}
Several extensions follow naturally. \emph{Beyond physical time.} More broadly, the principle extends to settings where $\Delta_k$ encodes any meaningful continuous parameter along which the sequence evolves, not necessarily wall-clock time. For example, trajectory modeling parameterized by arc length, or to any sequence indexed by a continuum the model can be made aware of analytically rather than as a learned feature. \emph{Active sampling.} A model with a faithful notion of its step parameter can be coupled to an acquisition rule that decides where the next observation should be taken. It is particularly relevant in clinical, scientific, and industrial settings where observations are costly. \emph{Robust Anytime forecasting.} Because $\Tilde{\Delta}$ is physical at inference, a trained TIDES can in principle forecast at arbitrary future steps without retraining by feeding the desired $\Delta$ at readout. Characterizing how this property degrades far from the training horizon distribution is a concrete open question.

\section{Conclusion}
\label{sec:conclusion}
The strengths of Mamba and S5, per-token expressivity and faithful physical-time semantics, are usually treated as belonging to different design philosophies. We show they are compatible once input-dependence is placed on the continuous-time generator rather than the discretization step: the model gains per-token expressivity, while the physical sampling interval keeps its physical role in the discretization. The resulting architecture, TIDES, achieves selective implicit time-awareness and sets new state-of-the-art results by highest accuracy on UEA classification and by average rank on Physiome-ODE forecasting, while matching or exceeding the reference baseline model on 6 of 8 natively irregular benchmarks. More broadly, our analysis suggests a design principle beyond state space models: keep expressivity on the continuous-time generator and time-handling analytic in the discretization.

\begin{ack}

The authors would like to thank Dr. Bekim Berisha (Inspire AG) for his support. No specific funding was received for this work.

\end{ack}

{
\small
\bibliographystyle{unsrtnat}
\bibliography{refs}

}

\appendix
\section{Architecture details}
\label{app:architecture}
\paragraph{Block execution order.}
Each TIDES block applies the following sequence of operations to its input $x \in \mathbb{R}^{B \times L \times H}$:
\begin{equation*}
  z = \mathrm{Dropout}\!\bigl(\mathrm{GLU}\!\bigl(\mathrm{Dropout}(\mathrm{GELU}(\mathrm{SSM}(\mathrm{BN}(x))))\bigr)\bigr) + x,
\end{equation*}
where BN is BatchNorm and both Dropout calls share the same rate \texttt{drop\_rate}.
The two Dropout applications are placed (i) immediately after the GELU nonlinearity, before the GLU, and (ii) after the GLU, before the residual addition.

\paragraph{BatchNorm without affine parameters.}
The BatchNorm layer is configured with no learnable scale or shift (\texttt{affine=False}), normalizing over the joint batch--time dimensions per channel.
A learnable affine rescaling would be redundant because the SSM recurrence and GLU provide all the learned scaling; more importantly, a per-channel scale that is constant across time would interfere with the HiPPO bias initialization.

\paragraph{Per-mode timescale initialization.}
Each of the $P$ complex modes has an independent learnable timescale parameter initialized uniformly in $[0.001, 0.1]$ range.
In practise, it is reparametrized in the log-space \.i.e. $\delta_p = \exp(\operatorname{log\_step}_p)$, which is positive by construction. It acts as a scaler per state dimension that is learned in the log-space.

\paragraph{Parallel scan.}
The associative scan uses the binary operator
$(A_i, b_i) \oplus (A_j, b_j) = (A_j A_i,\; b_j + A_j b_i)$,
applied in parallel over the sequence axis with $\mathcal{O}(\log L)$ depth and $\mathcal{O}(L)$ work per layer.
All operations are complex-valued on the diagonal state; the real and imaginary parts are stored as separate float channels to remain compatible with PyTorch autograd.

\paragraph{Bidirectional scan.}
When \texttt{bidir} is enabled, we run an additional pass over the reversed
input sequence and concatenate the forward and backward hidden states along
the channel dimension, treating the result as the hidden state.

\paragraph{GLU feed-forward expansion.}
The \texttt{ff\_mult} hyperparameter sets the expansion factor of the inner
dimension in the GLU feed-forward block (i.e.\ inner width
$= \texttt{ff\_mult} \cdot \texttt{hidden\_size}$).

\paragraph{Eigenvalue clipping.}
When \texttt{clip\_eigs} is enabled, we clamp the real part of the
continuous-time eigenvalues $\lambda$ to be at most $-10^{-5}$, preventing
eigenvalues from drifting into the right half-plane and ensuring stable
state transitions.

\paragraph{LR factor.}
LTI SSM matrices are commonly assigned a lower learning rate than the rest
of the network for stability reasons; the non-SSM learning rate is computed
as $\texttt{ssm\_lr} \cdot \texttt{lr\_factor}$. For TIDES, this lower rate
applies to the imaginary part of $\Lambda$, since it is the only SSM
parameter that remains LTI.

\paragraph{SSM block grouping.}
The \texttt{ssm\_b} hyperparameter partitions the diagonal state into independent groups, analogous to attention heads: the eigenvalues $\Lambda$ are split into \texttt{ssm\_b} disjoint groups, each evolving as its own SSM with its own $\Lambda$, $B$, and $C$ parameters.
The total state dimension is $\texttt{ssm\_size} = \texttt{ssm\_b} \cdot \texttt{ssm\_mult}$, where \texttt{ssm\_mult} sets the per-group state size.
This block structure trades off representational capacity (more groups capture more distinct dynamical modes) against per-group expressiveness (larger \texttt{ssm\_mult} allows richer dynamics within each group).

\section{Discretization derivations}
\label{app:disc}

We consider the continuous time linear state space model
\begin{equation}
    \dot{x}(t) = \Lambda\, x(t) + B\, u(t), \qquad y(t) = C\, x(t) + D\, u(t),
    \label{eq:cont_ssm}
\end{equation}
with diagonal state matrix $\Lambda \in \mathbb{C}^{P \times P}$, input matrix $B \in \mathbb{C}^{P \times H}$, and output matrix $C \in \mathbb{C}^{P \times P}$. To process discrete sequences sampled at (possibly irregular) timesteps $\{t_k\}_{k=0}^{L}$, we discretize \eqref{eq:cont_ssm} over each interval of length $\Delta_k = t_{k+1} - t_k$, yielding the recurrence
\begin{equation}
    x_{k+1} = \bar{\Lambda}_k\, x_k + \bar{B}_k\, u_k, \qquad y_k = C\, x_k + D\, u_k.
\end{equation}

\subsection{Zero order hold (ZOH)}
\label{app:disc:zoh}

Assuming the input $u(t)$ is held constant on $[t_k, t_{k+1})$ at value $u_k$, the exact solution of \eqref{eq:cont_ssm} is
\begin{equation}
    x(t_{k+1}) = e^{\Lambda \Delta_k}\, x(t_k) + \left( \int_{0}^{\Delta_k} e^{\Lambda (\Delta_k - \tau)} \mathrm{d}\tau \right) B\, u_k.
\end{equation}
For diagonal $\Lambda$ with strictly negative real part, $\Lambda$ is invertible and the integral admits the closed form
\begin{equation}
    \int_{0}^{\Delta_k} e^{\Lambda (\Delta_k - \tau)} \mathrm{d}\tau \;=\; \Lambda^{-1} \left( e^{\Lambda \Delta_k} - I \right).
\end{equation}
The ZOH discretized matrices are therefore
\begin{equation}
    \bar{\Lambda}_k^{\text{ZOH}} = \exp(\Lambda \Delta_k), \qquad
    \bar{B}_k^{\text{ZOH}} = \Lambda^{-1}\left( \exp(\Lambda \Delta_k) - I \right) B.
    \label{eq:zoh}
\end{equation}
Because $\Lambda$ is diagonal, the matrix exponential reduces to elementwise scalar exponentials, making \eqref{eq:zoh} cheap to compute and compatible with parallel scan.

\section{Mamba family}
\label{app:mamba}

Mamba~\citep{gu2023mamba}, Mamba-2~\citep{dao2024transformers} and Mamba-3~\citep{mamba3} differ in their state matrix and discretization rule. They share the property relevant to this paper: the step $\tilde{\Delta}_k$ is computed from the current token rather than set by the observation timestamps. Below, each recurrence is written for a single channel and indexed so that $x_k$ already contains $u_k$.

\paragraph{Mamba.}
The state matrix is real, diagonal and static, $A=\mathrm{diag}(a_1,\ldots,a_P)$ with $a_p<0$, while the step and projections are input-dependent:
\begin{align}
\tilde{\Delta}_k &= \mathrm{softplus}(W_{\Delta}u_k+b_{\Delta}), \qquad B_k=W_B u_k, \qquad C_k=W_C u_k, \label{eq:app_mamba_proj}\\
x_k &= e^{\tilde{\Delta}_k A}\,x_{k-1}+\tilde{\Delta}_k B_k\,u_k, \qquad y_k=C_k^{\top}x_k+D\,u_k. \label{eq:app_mamba1}
\end{align}
Although the original paper states a ZOH discretization, the implementation uses $\tilde{\Delta}_k B_k$ as the input matrix, a rule that \citet{mamba3} formalize as \emph{exponential Euler}. A short causal convolution precedes the SSM.

\paragraph{Mamba-2.}
Mamba-2 keeps Eq.~\eqref{eq:app_mamba1} and the step of Eq.~\eqref{eq:app_mamba_proj}, but restricts the state matrix to a static scalar per head, $A=a\,I_P$, with one $\tilde{\Delta}_k$ shared by all channels of the head. This enables the chunked matrix multiplication algorithm of state space duality (SSD) and larger state sizes, at the cost of a single decay rate per head. The changes target hardware efficiency; the handling of time is identical to Mamba.

\paragraph{Mamba-3.}
Mamba-3 modifies Mamba-2 in two ways relevant here. First, it replaces exponential Euler by an \emph{exponential trapezoidal} rule, which weights both interval endpoints:
\begin{equation}
x_k = e^{\tilde{\Delta}_k A_k}\,x_{k-1}+(1-\lambda_k)\,\tilde{\Delta}_k\,e^{\tilde{\Delta}_k A_k}B_{k-1}u_{k-1}+\lambda_k\,\tilde{\Delta}_k B_k u_k,
\label{eq:app_mamba3}
\end{equation}
where $\lambda_k\in[0,1]$ is computed from the token and $\lambda_k=1$ recovers Mamba-2. Second, the transition becomes complex, $e^{\tilde{\Delta}_k(A_k+i\theta_k)}$, with both the decay $A_k$ and the frequencies $\theta_k$ input-dependent; this is equivalent to a data-dependent rotary embedding on $B_k$ and $C_k$. Mamba-3 further adds a multi-input multi-output (MIMO) update for decoding efficiency and drops the short convolution.

\paragraph{What the family shares.}
In all three models, time enters the recurrence only through $\tilde{\Delta}_k=\mathrm{softplus}(W_\Delta u_k+b_\Delta)$. Under irregular sampling, $\Delta_k$ must therefore be appended to the input, $\tilde{\Delta}_k=\mathrm{softplus}(W_\Delta[u_k;\Delta_k]+b_\Delta)$ (Fig.~\ref{fig:high_level_s5_mamba_tides}), and the response of $\tilde{\Delta}_k$ to $\Delta_k$ must be learned from data. Nothing constrains this map to remain proportional to $\Delta_k$ outside the training range.

\paragraph{Real versus complex state.}
Mamba and Mamba-2 use real-valued state matrices, both in the official implementations and in our surrogate Mamba$_{\text{S}}$, whereas S5 and TIDES use a complex diagonal $\Lambda$. Mamba-3 is the only member of the family with complex dynamics, and there $\mathrm{Im}(\Lambda)$ is input-dependent.

\paragraph{Baselines in our experiments.}
Mamba$_{\text{S}}$ is the SSM core with a real diagonal state matrix and input-dependent $\tilde{\Delta}_k$, $B_k$, $C_k$, without convolution or gating, which isolates the placement of input-dependence at matched parameter count. Mamba, Mamba-2 and Mamba-3 use the official implementations, with Mamba-3 in its single-input single-output (SISO) configuration.
In all cases, $\Delta_k$ is provided as an extra input channel. We do not rescale $\tilde{\Delta}_k$ at inference by the ratio of training to test sampling rates, since this presupposes regular grids that differ by a single constant factor, which does not exist under irregular sampling.

\section{The \textit{Fading Flash} experiment: full details}
\label{app:toy}

\subsection{Data generation}

A sequence has length $L = 40$ and consists of two fields: a binary \emph{flash} indicator $p_k \in \{0,1\}$ and an integer \emph{zone index} $z_k \in \{0,1,2\}$ for each position $k=1,\ldots,L$.

\paragraph{Zone layout.}
Each sequence is assigned 2 or 3 contiguous zones (sampled uniformly).
Zone boundaries are drawn without replacement from positions $\{4,\ldots,35\}$, so every zone spans at least 4 positions.
Consecutive zones are guaranteed to have distinct rate indices.
Each zone $i$ is assigned a rate $\lambda_i \in \{1.0, 1.5, 2.0\}$ (slow / medium / fast) sampled uniformly, with the constraint that no two adjacent zones share a rate.

\paragraph{Flash positions.}
Between 2 and 4 flash positions are drawn without replacement from $\{0,\ldots,39\}$ and set to 1; all other positions are 0.

\paragraph{Target signal.}
Given a global timestep $\Delta$ and the per-position rate $\lambda_k = \lambda_{z_k}$, the glow $y_k$ evolves as
\begin{equation}
\label{eq:fading-flash}
    h_k = \alpha_k h_{k-1} + \beta_k p_k, \qquad y_k = h_k,
    \qquad \alpha_k = e^{-\lambda_k \Delta}, \quad \beta_k = \tfrac{1 - \alpha_k}{\lambda_k}.
\end{equation}
This is the exact zero-order-hold discretization of a continuous first-order system $\dot{h} = -\lambda_k h + p$, sampled at interval $\Delta$.

\paragraph{Model input.}
The input to each model at position $k$ is $u_k \in \mathbb{R}^4$: the flash indicator $p_k$ concatenated with the one-hot encoding of the zone $z_k \in \{0,1,2\}$.
The zone identity is thus explicit in the input; the challenge is to use it to select the correct decay rate, then generalize that selection across unseen $\Delta$ values.

\paragraph{Training distribution.}
During training, $\Delta$ is sampled uniformly from $[0.5, 1.5]$ for each batch element independently.

\subsection{Model details}

All models share the same input and output interface: they receive the sequence $u \in \mathbb{R}^{L \times 4}$ and the scalar $\Delta$, and predict $\hat{y} \in \mathbb{R}^{L \times 1}$.
The architectures use a simplified, self-contained SSM without the full block structure (no BatchNorm, no GLU, no dropout), isolating the SSM core.

Each consists of a linear encoder $W_{\mathrm{enc}} : \mathbb{R}^4 \to \mathbb{R}^H$ followed by a real-diagonal SSM with $P$ states and a linear readout $C \in \mathbb{R}^{1 \times P}$, and a direct feedthrough $D \in \mathbb{R}^{1 \times H}$.
The state update uses ZOH discretization.
The three variants differ only in which parameters are input-dependent: for S5 and TIDES variants, $\Tilde{\Delta}_k$ is the physical timestep, i.e., $\Tilde{\Delta}_k=\Delta_k$, and enters through multiplication, not through the network; for the Mamba surrogate, the step is $\mathrm{softplus}(W_{\Delta} [h_k, \Delta_k])$, where $h_k = W_{\mathrm{enc}} u_k$, so $\tilde\Delta_k$ is learned as a function of the input.

\subsection{Training protocol}

All models are trained with Adam ($\mathrm{lr} = 3 \times 10^{-3}$, default $\beta_1,\beta_2$) for 3000 steps with batch size 32, minimizing MSE against the ground-truth glow \eqref{eq:fading-flash}.
All runs use seed 0.
No learning-rate schedule or weight decay is applied.

\subsection{Evaluation protocol}

Models are evaluated at ten test values $\Delta \in \{0.1, 0.2, 0.3, 0.5, 0.8, 1.0, 1.2, 1.5, 1.8, 2.0\}$.
The training range $\Delta \in [0.5, 1.5]$ covers positions 4–8 in this grid; positions 1–3 and 9–10 are out-of-distribution.
At each $\Delta$, we draw $6 \times 64 = 384$ fresh sequences at that fixed $\Delta$ and report the mean MSE.

The primary metric is \emph{relative error} $= \sqrt{\mathrm{MSE} / \mathrm{Var}(y)} \times 100\%$, where $\mathrm{Var}(y)$ is estimated from $10 \times 128 = 1280$ samples drawn at the same $\Delta$.
Normalizing by target variance makes the metric comparable across $\Delta$ values: a target with small $\Delta$ has small glow values, and raw MSE would be misleadingly low.

\subsection{Results}

\begin{figure}
    \centering                           
    \includegraphics[width=0.7\linewidth]{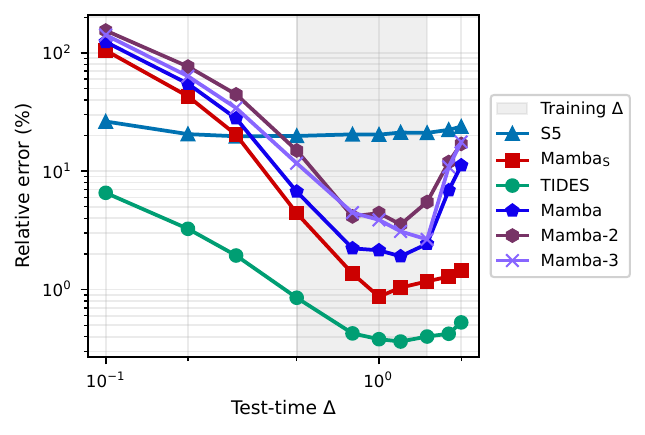}        
    \caption{\textbf{Relative error vs.\ test $\Delta$.} Mamba family consistently under-performs TIDES.}                                  
    \label{fig:fading_flash_with_mamba}
\end{figure}

At matched parameter count, the Mamba variants do not extrapolate to unseen step sizes (Fig.~\ref{fig:fading_flash_with_mamba}). At $\Delta$ = 0.1, the relative error is 6.56 for TIDES against 141.20 for Mamba-3 and 154.08 for Mamba-2, and the ordering holds across the whole sweep. Mamba-3 improves on Mamba-2 but does not change the failure mode, which is the point we make in the paper: the improvements of the Mamba family are orthogonal to where input-dependence is placed.

\section{Dataset details and experimental setup}
\label{app:experiments}

We note that the community has recently concluded that the previous irregularly sampled multivariate time series (IMTS) forecasting benchmarks (PhysioNet2012 \cite{pollard_physionet_2026}, MIMIC-III/IV \cite{johnson2016mimic,johnson2023mimiciv}) are poor instruments for measuring temporal modeling. These datasets feature a time-blind constant-prediction baseline without access to the query timepoints; they originate as mortality-classification challenges repurposed for forecasting; and required covariates are often absent, resulting in ill-posed tasks \cite{klotergens2025physiome}. Physiome-ODE solves these deficiencies by spanning 50 distinct biophysical systems, guaranteeing that the generating covariates are present, and separating methods rather than collapsing them into ties \cite{klotergens2025physiome}. It is the \textit{de facto} benchmark for IMTS forecasting, hence our evaluation on it. On the classification side, natively irregular corpora are scarce, so the field simulates irregularity by dropping observations from regular grids, discarding the structural, non-random missingness that real/experimental data collection would produce. 

\subsection{UEA classification (Walker 2024 protocol)}
\label{app:uea}

We use six datasets from the UEA multivariate time series classification archive \citep{bagnall2018uea}: SelfRegulationSCP1 (SCP1), SelfRegulationSCP2 (SCP2), MotorImagery (MI), EigenWorms (EW), EthanolConcentration (ETC), and Heartbeat (HB). These are the same six long-context classification problems used in the LogNCDE evaluation \cite{walker2024log} and reproduced by RFormer \citep{moreno2024rough}, selected to span EEG/MEG (SCP1, SCP2, MI, HB), motion-capture worm dynamics (EW), and spectroscopy (ETC). Sequence lengths range from $L{=}405$ (HB) to $L{=}17{,}984$ (EW); channel counts range from $3$ (ETC) to $64$ (MI). All series in each problem are equal length and contain no missing values; we use the archive's pre-stored ARFF files without further interpolation. Per-channel inputs are standardized to zero mean and unit variance using statistics computed on the training partition only.

\paragraph{Splits and task.} We follow the RFormer protocol bit-for-bit. The original UEA train and test partitions are concatenated and re-split into train / validation / test using a $70/15/15$ random partition with five seeds; all reported numbers are mean $\pm$ std over the five resulting folds. The task is closed-set multivariate classification, evaluated by accuracy on the held-out test partition. No data augmentation, time warping, or downsampling is applied (Tables~\ref{tab:uea_winner_configs} and~\ref{tab:search_grid}).

\begin{table}[h!]
\centering
\caption{Best TIDES configurations per UEA dataset.}
\label{tab:uea_winner_configs}
\resizebox{\linewidth}{!}{%
\begin{tabular}{l r r r r r r r l l r r r r l l r}
\toprule
\textbf{Dataset}
  & \textbf{lr}
  & \textbf{wd}
  & \textbf{$h$}
  & \textbf{ssm}
  & \textbf{ssm\_b}
  & \textbf{$L$}
  & \textbf{$d_{\text{enc}}$}
  & \textbf{learn\,$\lambda$}
  & \textbf{disc}
  & \textbf{drop}
  & \textbf{bs}
  & \textbf{$d_{\lambda}$}
  & \textbf{bc\_rank}
  & \textbf{bidir}
  & \textbf{clip}
  & \textbf{ep} \\
\midrule
SCP1
  & $2.17{\times}10^{-4}$ & $0.0$
  & $16$ & $32$ & $4$ & $4$ & $0$
  & exp & zoh
  & $0.15$ & $5$ & $0$ & $8$
  & $-$ & $-$ & $200$ \\

SCP2
  & $2.72{\times}10^{-4}$ & $0.0$
  & $16$ & $64$ & $8$ & $2$ & $0$
  & stable & zoh
  & $0.00$ & $10$ & $0$ & $16$
  & \checkmark & \checkmark & $400$ \\

MI
  & $6.52{\times}10^{-4}$ & $0.01$
  & $32$ & $16$ & $4$ & $4$ & $0$
  & standard & zoh
  & $0.05$ & $20$ & $0$ & $16$
  & \checkmark & \checkmark & $200$ \\

EW
  & $1.56{\times}10^{-3}$ & $0.1$
  & $16$ & $16$ & $2$ & $1$ & $0$
  & stable & zoh
  & $0.00$ & $10$ & $0$ & $16$
  & \checkmark & \checkmark & $400$ \\

ETC
  & $5.18{\times}10^{-4}$ & $0.1$
  & $8$ & $32$ & $8$ & $8$ & $1$
  & exp & zoh
  & $0.05$ & $20$ & $2$ & $8$
  & \checkmark & $-$ & $200$ \\

HB
  & $3.32{\times}10^{-4}$ & $0.1$
  & $64$ & $16$ & $4$ & $4$ & $0$
  & exp & zoh
  & $0.10$ & $20$ & $0$ & $8$
  & \checkmark & $-$ & $400$ \\
\bottomrule
\end{tabular}%
}
\end{table}

\begin{table}[h!]
\centering
\caption{Hyperparameter search space for TIDES on the UEA benchmark.
  Continuous parameters use log-uniform sampling; all others are categorical.}
\label{tab:search_grid}
\begin{tabular}{l l}
\toprule
\textbf{Hyperparameter} & \textbf{Values / Range} \\
\midrule
lr & $[10^{-6},\; 10^{-2}]$ \\
weight decay & $\{0,\; 10^{-3},\; 10^{-2},\; 10^{-1}\}$ \\
$h$ (hidden dim) & $\{8,\; 16,\; 32,\; 64\}$ \\
$L$ (layers) & $\{1,\; 2,\; 4,\; 6,\; 8\}$ \\
ssm\_b (SSM blocks) & $\{2,\; 4,\; 8\}$ \\
ssm\_mult & $\{1,\; 2,\; 4\}$ \\
disc & \{zoh,\; bilinear\} \\
drop & $\{0.00,\; 0.05,\; 0.10,\; 0.15,\; 0.20\}$ \\
$\lambda$ reparametrization & \{standard,\; stable,\; exp\} \\
bidir & \{True,\; False\} \\
$d_{\text{enc}}$ & $\{0,\; 1,\; 2\}$ \\
$d_{\lambda}$ & $\{0,\; 1,\; 2\}$ \\
bc\_rank & $\{4,\; 8,\; 16\}$ \\
batch size & $\{5,\; 10,\; 20\}$ \\
epochs & $\{200,\; 400\}$ \\
clip\_eigs & \{True,\; False\} \\
\bottomrule
\end{tabular}
\end{table}

\subsection{Physiome-ODE (Kl\"otergens 2025 protocol)}
\label{app:physiome-data}

We use the public Physiome-ODE benchmark \citep{klotergens2025physiome} without modification. Physiome-ODE comprises $50$ irregularly-sampled multivariate time-series (IMTS) forecasting datasets, each generated by simulating a biological ordinary differential equation; the $50$ ODEs were selected from a pool of $208$ candidate models by ranking on the Joint Gradient Deviation (JGD) score \cite{klotergens2025physiome}, which jointly captures within-trajectory gradient variance and across-trajectory diversity. For each ODE, the authors stochastically perturb the literature initial conditions, ODE constants, and integration duration, with spreads $(\sigma_{\mathrm{initial}}, \sigma_{\mathrm{const}}, \sigma_{\mathrm{dur}})$ optimized to maximize JGD over $\sigma_{\mathrm{initial}} \in \{0.1, 0.3, 0.5\}$, $\sigma_{\mathrm{const}} \in \{0.05, 0.1, 0.3\}$, and $\sigma_{\mathrm{dur}} \in \{0.33, 1, 3.3, 10, 30\}$. Each dataset contains $2{,}000$ trajectories, each with $200$ regularly-spaced ODE solutions over duration $\tfrac{1}{2}\sigma^*_{\mathrm{dur}}$. To turn these into IMTS instances, $80\%$ of observations are randomly masked out per channel and additive Gaussian noise $\varepsilon \sim \mathcal{N}(0, 0.05)$ is applied to the retained values. All channels are normalized to zero mean and unit variance over the union of all timesteps and instances within a dataset. For multivariate ODEs, the JGD score reported in Table \ref{tab:results} is the mean of the ten channels with the highest per-channel JGD (as per the protocol of Kl\"otergens et al. \cite{klotergens2025physiome}, designed to avoid diluting JGD by trivially-constant channels).

\paragraph{Splits and task.} We use the benchmark's published $5$-fold cross-validation with a $70/20/10$ train/validation/test instance-level split per fold; the observation mask is resampled per instance per fold to maintain the $80\%$ sparsity. The forecasting target is the second half of each trajectory ($100$ steps) given the first half ($100$ steps) as conditioning. Performance is reported as masked MSE on observed entries in the forecast window, averaged over the five folds; per-dataset means and standard deviations appear in Table~\ref{tab:results}.

\paragraph{Training.} All models are trained for $200$ epochs with early stopping on validation MSE (patience $30$), using AdamW with a cosine-annealed learning rate and linear warmup over the first \texttt{warmup\_epochs} epochs. The training objective is mean squared error on observed entries in the forecast window. Following the protocol of Kl\"otergens et al. \cite{klotergens2025physiome}, we sample $10$ hyperparameter configurations and select the best on the first fold's validation set, then evaluate the selected configuration on all five folds; reported numbers in Table \ref{tab:results} are mean $\pm$ std across folds. The hyperparameter search space is given in Table \ref{tab:hp-physiome}.

\begin{table}[h]
\centering
\caption{Hyperparameter search space for Physiome-ODE.
  Continuous parameters use log-uniform sampling; all others are categorical.}
\label{tab:hp-physiome}
\begin{tabular}{l l}
\toprule
\textbf{Hyperparameter} & \textbf{Values / Range} \\
\midrule
lr & $[10^{-6},\; 10^{-3}]$ \\
lr\_factor & $[1,\; 350]$ \\
weight\_decay & $[10^{-6},\; 10^{-1}]$ \\
warmup\_epochs & $\{1,\; 5,\; 10\}$ \\
batch\_size & $\{16,\; 32,\; 48,\; \ldots,\; 128\}$ \\
drop\_rate & $\{0.00,\; 0.05,\; 0.10,\; \ldots,\; 0.30\}$ \\
$h$ (hidden\_size) & $\{16,\; 20,\; 24,\; \ldots,\; 192\}$ \\
num\_blocks & $[2,\; 16]$ \\
$d_{\text{enc}}$ (encoder\_depth) & $\{0,\; 1,\; 2\}$ \\
ff\_mult & $\{0.5\} \cup \{1,\; 2,\; \ldots,\; 8\}$ \\
bidir & \{True,\; False\} \\
ssm\_size & $\{2,\; 4,\; 6,\; \ldots,\; 50\}$ \\
bc\_rank & $[1,\; 48]$ \\
$d_{\lambda}$ (lambda\_encoder\_depth) & $\{0,\; 1,\; 2\}$ \\
$\lambda$ reparametrization & \{standard,\; stable,\; exp,\; softplus\} \\
disc (discretization) & \{zoh,\; bilinear\} \\
clip\_eigs & \{True,\; False\} \\
proj\_init\_method & \{zeros,\; random\} \\
\bottomrule
\end{tabular}
\end{table}

\section{UEA complete results}
\label{app:uea_results}

\paragraph{Results.} Herein we report the complete results for time series classification on the UEA dataset, also including the Transformer, NCDE, NRDE, Mamba-2, and Mamba models (Table \ref{tab:uea_full_results}).

\begin{table}[h!]
\centering
\caption{UEA time series classification results (accuracy \% $\pm$ standard deviation). Bold indicates the best result per dataset. Baseline accuracies are reproduced from Walker et al. \cite{walker2024log}, Moreno et al. \cite{moreno2024rough}, and Mamba-2 from \cite{saadatmand2026simple}.}
\label{tab:uea_full_results}
\footnotesize
\setlength{\tabcolsep}{4pt}
\resizebox{\linewidth}{!}{%
\begin{tabular}{lcccccccc}
\toprule
Model & SCP1 & SCP2 & MI & EW & ETC & HB & Avg & Avg Rank\\
\midrule
\textbf{TIDES}     & $87.6{\pm}3.3$ & $\mathbf{56.5{\pm}6.2}$ & $\mathbf{58.2{\pm}2.8}$ & $85.7{\pm}4.0$ & $34.2{\pm}2.4$ & $74.4{\pm}4.0$ & $\mathbf{66.1}$ & $\mathbf{3.00}$ \\
Mamba-3     & $83.3{\pm}2.8$ & $55.4{\pm}3.6$ & $57.9{\pm}5.4$ & $84.0{\pm}2.6$ & $34.9{\pm}4.8$ & $72.8{\pm}4.3$ & 64.7 & 4.17 \\
LogNCDE     & $83.1{\pm}2.8$ & $53.7{\pm}4.1$ & $53.7{\pm}5.3$ & $85.6{\pm}5.1$ & $34.4{\pm}6.4$ & $75.2{\pm}4.6$ & 64.3 & 4.25 \\
RFormer     & $81.2{\pm}2.8$ & $52.3{\pm}3.7$ & $55.8{\pm}6.6$ & $\mathbf{90.3{\pm}0.1}$ & $34.7{\pm}4.1$ & $72.5{\pm}0.1$ & 64.5 & 5.17 \\
S6          & $82.8{\pm}2.7$ & $49.9{\pm}9.5$ & $51.3{\pm}4.7$ & $85.0{\pm}16.1$ & $26.4{\pm}6.4$ & $76.5{\pm}8.3$ & 62.0 & 6.33 \\
LRU         & $82.6{\pm}3.4$ & $51.2{\pm}3.6$ & $48.4{\pm}5.0$ & $87.8{\pm}2.8$ & $21.5{\pm}2.1$ & $\mathbf{78.4{\pm}6.7}$ & 61.7 & 6.50 \\
S5          & $\mathbf{89.9{\pm}4.6}$ & $50.5{\pm}2.6$ & $47.7{\pm}5.5$ & $81.1{\pm}3.7$ & $24.1{\pm}4.3$ & $77.7{\pm}5.5$ & 61.8 & 7.08 \\
Transformer & $84.3{\pm}6.3$ & $49.1{\pm}2.5$ & $50.5{\pm}3.0$ & OOM & $\mathbf{40.5{\pm}6.3}$ & $70.5{\pm}0.1$ & 59.0 & 7.50 \\
NCDE        & $79.8{\pm}5.6$ & $53.0{\pm}2.8$ & $49.5{\pm}2.8$ & $75.0{\pm}3.9$ & $29.9{\pm}6.5$ & $73.9{\pm}2.6$ & 60.2 & 7.83 \\
NRDE        & $80.9{\pm}2.5$ & $53.7{\pm}6.9$ & $47.0{\pm}5.7$ & $83.9{\pm}7.3$ & $25.3{\pm}1.8$ & $72.9{\pm}4.8$ & 60.6 & 8.25 \\
Mamba-2     & $78.5{\pm}3.3$ & $51.5{\pm}2.8$ & $50.4{\pm}3.3$ & $83.4{\pm}3.8$ & $28.9{\pm}2.3$ & $72.4{\pm}2.6$ & 60.9 & 8.67 \\
Mamba       & $80.7{\pm}1.4$ & $48.2{\pm}3.9$ & $47.7{\pm}4.5$ & $70.9{\pm}15.8$ & $27.9{\pm}4.5$ & $76.2{\pm}3.8$ & 58.6 & 9.25 \\
\bottomrule
\end{tabular}%
}
\end{table}

\section{Physiome-ODE per-dataset results}
\label{app:physiome}

Table~\ref{tab:results} reports the full per-dataset results across all 50 Physiome-ODE problems, ordered by descending Joint Gradient Deviation (JGD). For each dataset, we report the $5$-fold mean and standard deviation of the masked forecasting MSE, with the best score per row in \textbf{bold}. Aggregate statistics over the suite are reported in the bottom two rows and include the number of best-method wins per model and the average rank, which depends on the baseline pool and on reproduced numbers \cite{walker2024log, moreno2024rough, klotergens2025physiome}. TIDES achieves the best average rank (2.40) and ties for the most wins (16 of 50, jointly with LinODEnet); the remaining wins are distributed across GraFITi, GraFITi-C, and CRU, with no architecture dominating across the JGD spectrum. The two extremes of the table behave differently: high-JGD problems (top) are dominated by the constant-channel baseline GraFITi-C, indicating that several of the most ``volatile'' ODEs offer little forecastable signal beyond their channel mean, whereas the low-JGD regime (bottom) cleanly separates the methods, with LinODEnet and TIDES alternating in first place.

\begin{table}[h!]
\centering
\caption{Physiome-ODE benchmark results (MSE). Bold indicates the best result per dataset. Baseline MSEs are reproduced from \citet{klotergens2025physiome}; TIDES is our run.}
\label{tab:results}
\footnotesize
\resizebox{\linewidth}{!}{%
\begin{tabular}{lrccccccc}
\toprule
Dataset & JGD & TIDES & LinODEnet & GraFITi & GraFITi-C & CRU & Neural Flows & GRU-ODE-Bayes \\
\midrule
DUP01 & 2.697 & $0.953{\pm}0.037$ & $0.964{\pm}0.036$ & $0.955{\pm}0.037$ & $\mathbf{0.951{\pm}0.036}$ & $0.958{\pm}0.038$ & $1.030{\pm}0.046$ & $1.037{\pm}0.047$ \\
JEL01 & 2.580 & $0.937{\pm}0.014$ & $0.949{\pm}0.015$ & $0.942{\pm}0.020$ & $\mathbf{0.935{\pm}0.016}$ & $0.939{\pm}0.015$ & $1.000{\pm}0.013$ & $1.017{\pm}0.014$ \\
DOK01 & 2.277 & $\mathbf{0.982{\pm}0.004}$ & $0.996{\pm}0.003$ & $0.984{\pm}0.005$ & $0.982{\pm}0.005$ & $0.985{\pm}0.005$ & $0.998{\pm}0.005$ & $1.011{\pm}0.004$ \\
INA01 & 2.218 & $\mathbf{1.003{\pm}0.010}$ & $1.009{\pm}0.011$ & $1.004{\pm}0.010$ & $1.004{\pm}0.009$ & $1.005{\pm}0.010$ & $1.008{\pm}0.008$ & $1.018{\pm}0.010$ \\
WOL01 & 1.973 & $0.807{\pm}0.027$ & $0.806{\pm}0.027$ & $0.787{\pm}0.028$ & $\mathbf{0.784{\pm}0.030}$ & $0.814{\pm}0.029$ & $0.841{\pm}0.029$ & $0.952{\pm}0.035$ \\
BOR01 & 1.795 & $0.711{\pm}0.022$ & $0.719{\pm}0.021$ & $0.712{\pm}0.022$ & $\mathbf{0.709{\pm}0.022}$ & $0.715{\pm}0.020$ & $0.743{\pm}0.026$ & $0.794{\pm}0.034$ \\
HYN01 & 1.548 & $1.026{\pm}0.068$ & $0.672{\pm}0.044$ & $0.625{\pm}0.043$ & $\mathbf{0.619{\pm}0.046}$ & $0.665{\pm}0.053$ & $0.636{\pm}0.045$ & $0.883{\pm}0.060$ \\
JEL02 & 1.271 & $0.777{\pm}0.032$ & $0.693{\pm}0.031$ & $0.699{\pm}0.029$ & $0.687{\pm}0.027$ & $\mathbf{0.674{\pm}0.028}$ & $0.779{\pm}0.028$ & $0.816{\pm}0.049$ \\
DUP02 & 1.202 & $0.722{\pm}0.044$ & $0.740{\pm}0.042$ & $0.728{\pm}0.044$ & $\mathbf{0.718{\pm}0.046}$ & $0.722{\pm}0.046$ & $0.890{\pm}0.081$ & $0.895{\pm}0.055$ \\
WOL02 & 0.895 & $0.650{\pm}0.015$ & $0.663{\pm}0.015$ & $0.654{\pm}0.014$ & $\mathbf{0.645{\pm}0.016}$ & $0.653{\pm}0.017$ & $0.685{\pm}0.012$ & $0.854{\pm}0.010$ \\
DIF01 & 0.735 & $\mathbf{0.706{\pm}0.054}$ & $0.832{\pm}0.087$ & $0.985{\pm}0.030$ & $0.982{\pm}0.029$ & $0.985{\pm}0.025$ & $1.014{\pm}0.025$ & $1.035{\pm}0.023$ \\
VAN01 & 0.407 & $0.249{\pm}0.005$ & $0.250{\pm}0.006$ & $0.246{\pm}0.005$ & $\mathbf{0.242{\pm}0.006}$ & $0.253{\pm}0.005$ & $0.250{\pm}0.006$ & $0.321{\pm}0.023$ \\
DUP03 & 0.254 & $\mathbf{0.621{\pm}0.047}$ & $0.632{\pm}0.044$ & $0.627{\pm}0.043$ & $0.744{\pm}0.042$ & $0.622{\pm}0.047$ & $1.098{\pm}0.447$ & $0.874{\pm}0.089$ \\
BER01 & 0.179 & $\mathbf{0.265{\pm}0.016}$ & $0.279{\pm}0.020$ & $0.300{\pm}0.018$ & $0.342{\pm}0.018$ & $0.280{\pm}0.016$ & $0.398{\pm}0.014$ & $0.594{\pm}0.054$ \\
LEN01 & 0.178 & $\mathbf{0.383{\pm}0.023}$ & $0.387{\pm}0.071$ & $0.607{\pm}0.055$ & $0.970{\pm}0.063$ & $0.754{\pm}0.157$ & $1.009{\pm}0.061$ & $1.028{\pm}0.060$ \\
LI01 & 0.113 & $0.111{\pm}0.006$ & $\mathbf{0.084{\pm}0.009}$ & $0.202{\pm}0.013$ & $0.742{\pm}0.010$ & $0.175{\pm}0.020$ & $0.979{\pm}0.049$ & $1.067{\pm}0.015$ \\
LI02 & 0.097 & $\mathbf{0.380{\pm}0.049}$ & $0.434{\pm}0.044$ & $0.397{\pm}0.058$ & $0.458{\pm}0.056$ & $0.437{\pm}0.046$ & $0.674{\pm}0.105$ & $0.723{\pm}0.080$ \\
REV01 & 0.081 & $0.607{\pm}0.040$ & $\mathbf{0.597{\pm}0.061}$ & $0.674{\pm}0.055$ & $0.855{\pm}0.050$ & $0.602{\pm}0.049$ & $0.978{\pm}0.042$ & $1.091{\pm}0.075$ \\
PUR01 & 0.049 & $0.134{\pm}0.013$ & $\mathbf{0.106{\pm}0.006}$ & $0.153{\pm}0.006$ & $0.476{\pm}0.020$ & $0.353{\pm}0.083$ & $0.654{\pm}0.102$ & $0.703{\pm}0.058$ \\
NYG01 & 0.047 & $0.369{\pm}0.084$ & $0.358{\pm}0.071$ & $\mathbf{0.344{\pm}0.065}$ & $0.366{\pm}0.047$ & $0.403{\pm}0.092$ & $0.442{\pm}0.094$ & $0.571{\pm}0.074$ \\
PUR02 & 0.044 & $\mathbf{0.271{\pm}0.020}$ & $0.280{\pm}0.028$ & $0.322{\pm}0.021$ & $0.511{\pm}0.023$ & $0.293{\pm}0.026$ & $0.723{\pm}0.077$ & $0.830{\pm}0.066$ \\
HOD01 & 0.042 & $\mathbf{0.408{\pm}0.032}$ & $0.441{\pm}0.043$ & $0.493{\pm}0.046$ & $0.609{\pm}0.056$ & $0.409{\pm}0.049$ & $0.701{\pm}0.077$ & $0.851{\pm}0.118$ \\
REE01 & 0.035 & $\mathbf{0.022{\pm}0.002}$ & $0.045{\pm}0.012$ & $0.033{\pm}0.007$ & $0.039{\pm}0.012$ & $0.051{\pm}0.008$ & $0.045{\pm}0.011$ & $0.266{\pm}0.068$ \\
VIL01 & 0.028 & $0.355{\pm}0.021$ & $0.374{\pm}0.021$ & $\mathbf{0.344{\pm}0.044}$ & $0.378{\pm}0.042$ & $0.373{\pm}0.039$ & $0.500{\pm}0.060$ & $0.511{\pm}0.053$ \\
KAR01 & 0.023 & $0.035{\pm}0.007$ & $\mathbf{0.034{\pm}0.008}$ & $0.041{\pm}0.013$ & $0.078{\pm}0.011$ & $0.044{\pm}0.012$ & $0.069{\pm}0.009$ & $0.193{\pm}0.013$ \\
SHO01 & 0.023 & $0.093{\pm}0.019$ & $0.057{\pm}0.006$ & $0.062{\pm}0.013$ & $\mathbf{0.055{\pm}0.013}$ & $0.095{\pm}0.010$ & $0.092{\pm}0.014$ & $0.260{\pm}0.017$ \\
BUT01 & 0.020 & $\mathbf{0.253{\pm}0.067}$ & $0.254{\pm}0.074$ & $0.281{\pm}0.071$ & $0.324{\pm}0.091$ & $0.317{\pm}0.108$ & $0.441{\pm}0.153$ & $0.583{\pm}0.172$ \\
MAL01 & 0.019 & $0.041{\pm}0.003$ & $\mathbf{0.018{\pm}0.007}$ & $0.020{\pm}0.004$ & $0.054{\pm}0.005$ & $0.064{\pm}0.007$ & $0.052{\pm}0.002$ & $0.420{\pm}0.061$ \\
ASL01 & 0.019 & $0.035{\pm}0.004$ & $\mathbf{0.022{\pm}0.003}$ & $0.025{\pm}0.009$ & $0.026{\pm}0.002$ & $0.046{\pm}0.014$ & $0.066{\pm}0.033$ & $0.114{\pm}0.035$ \\
BUT02 & 0.016 & $\mathbf{0.205{\pm}0.029}$ & $0.207{\pm}0.056$ & $0.248{\pm}0.052$ & $0.256{\pm}0.039$ & $0.282{\pm}0.042$ & $0.329{\pm}0.068$ & $0.483{\pm}0.047$ \\
MIT01 & 0.015 & $\mathbf{0.003{\pm}0.000}$ & $0.003{\pm}0.000$ & $0.003{\pm}0.000$ & $0.003{\pm}0.000$ & $0.003{\pm}0.000$ & $0.008{\pm}0.004$ & $0.005{\pm}0.001$ \\
GUP01 & 0.014 & $0.041{\pm}0.017$ & $\mathbf{0.018{\pm}0.007}$ & $0.041{\pm}0.006$ & $0.035{\pm}0.006$ & $0.057{\pm}0.017$ & $0.043{\pm}0.017$ & $0.153{\pm}0.041$ \\
GUY01 & 0.013 & $0.004{\pm}0.001$ & $0.006{\pm}0.005$ & $0.005{\pm}0.003$ & $0.004{\pm}0.001$ & $\mathbf{0.004{\pm}0.001}$ & $0.024{\pm}0.015$ & $0.036{\pm}0.004$ \\
PHI01 & 0.013 & $\mathbf{0.128{\pm}0.022}$ & $0.131{\pm}0.014$ & $0.222{\pm}0.013$ & $0.345{\pm}0.015$ & $0.133{\pm}0.020$ & $0.635{\pm}0.158$ & $0.674{\pm}0.030$ \\
GUY02 & 0.013 & $\mathbf{0.009{\pm}0.001}$ & $0.010{\pm}0.006$ & $0.012{\pm}0.009$ & $0.032{\pm}0.015$ & $0.010{\pm}0.002$ & $0.082{\pm}0.066$ & $0.124{\pm}0.044$ \\
PUL01 & 0.013 & $0.009{\pm}0.002$ & $\mathbf{0.008{\pm}0.004}$ & $0.008{\pm}0.001$ & $0.024{\pm}0.008$ & $0.012{\pm}0.003$ & $0.061{\pm}0.060$ & $0.099{\pm}0.024$ \\
CAL01 & 0.013 & $0.103{\pm}0.012$ & $\mathbf{0.078{\pm}0.009}$ & $0.179{\pm}0.012$ & $0.643{\pm}0.024$ & $0.158{\pm}0.008$ & $0.867{\pm}0.014$ & $1.049{\pm}0.055$ \\
WOD01 & 0.012 & $0.120{\pm}0.011$ & $0.154{\pm}0.016$ & $0.164{\pm}0.013$ & $0.344{\pm}0.016$ & $\mathbf{0.113{\pm}0.017}$ & $0.510{\pm}0.060$ & $0.612{\pm}0.072$ \\
GUP02 & 0.012 & $\mathbf{0.314{\pm}0.018}$ & $0.469{\pm}0.022$ & $0.449{\pm}0.027$ & $0.461{\pm}0.025$ & $0.444{\pm}0.018$ & $0.577{\pm}0.017$ & $0.870{\pm}0.059$ \\
M01 & 0.012 & $0.004{\pm}0.001$ & $0.004{\pm}0.001$ & $\mathbf{0.003{\pm}0.000}$ & $0.003{\pm}0.000$ & $0.005{\pm}0.000$ & $0.124{\pm}0.207$ & $0.055{\pm}0.009$ \\
LEN02 & 0.012 & $0.041{\pm}0.003$ & $\mathbf{0.039{\pm}0.005}$ & $0.099{\pm}0.021$ & $0.143{\pm}0.022$ & $0.059{\pm}0.012$ & $0.380{\pm}0.141$ & $0.297{\pm}0.071$ \\
KAR02 & 0.011 & $0.159{\pm}0.007$ & $\mathbf{0.140{\pm}0.010}$ & $0.151{\pm}0.009$ & $0.252{\pm}0.010$ & $0.151{\pm}0.011$ & $0.257{\pm}0.016$ & $0.515{\pm}0.032$ \\
SHO02 & 0.011 & $0.062{\pm}0.016$ & $\mathbf{0.037{\pm}0.006}$ & $0.043{\pm}0.006$ & $0.073{\pm}0.010$ & $0.083{\pm}0.015$ & $0.109{\pm}0.015$ & $0.368{\pm}0.028$ \\
MAC01 & 0.010 & $0.047{\pm}0.003$ & $0.020{\pm}0.003$ & $0.021{\pm}0.003$ & $\mathbf{0.019{\pm}0.002}$ & $0.065{\pm}0.006$ & $0.029{\pm}0.011$ & $0.242{\pm}0.026$ \\
IRI01 & 0.010 & $0.047{\pm}0.012$ & $\mathbf{0.037{\pm}0.003}$ & $0.038{\pm}0.017$ & $0.097{\pm}0.008$ & $0.049{\pm}0.010$ & $0.116{\pm}0.006$ & $0.151{\pm}0.032$ \\
BAG01 & 0.010 & $0.037{\pm}0.002$ & $0.032{\pm}0.005$ & $\mathbf{0.029{\pm}0.002}$ & $0.109{\pm}0.002$ & $0.046{\pm}0.005$ & $0.075{\pm}0.012$ & $0.294{\pm}0.041$ \\
WOL03 & 0.008 & $0.125{\pm}0.010$ & $\mathbf{0.073{\pm}0.010}$ & $0.105{\pm}0.016$ & $0.247{\pm}0.032$ & $0.177{\pm}0.016$ & $0.479{\pm}0.107$ & $0.859{\pm}0.114$ \\
WAN01 & 0.008 & $0.115{\pm}0.015$ & $\mathbf{0.103{\pm}0.012}$ & $0.119{\pm}0.010$ & $0.232{\pm}0.015$ & $0.125{\pm}0.012$ & $0.345{\pm}0.027$ & $0.504{\pm}0.031$ \\
NEL01 & 0.007 & $0.009{\pm}0.000$ & $0.010{\pm}0.001$ & $\mathbf{0.007{\pm}0.000}$ & $0.023{\pm}0.006$ & $0.009{\pm}0.001$ & $0.060{\pm}0.029$ & $0.054{\pm}0.012$ \\
HUA01 & 0.007 & $0.087{\pm}0.009$ & $\mathbf{0.052{\pm}0.004}$ & $0.063{\pm}0.005$ & $0.115{\pm}0.007$ & $0.116{\pm}0.007$ & $0.149{\pm}0.015$ & $0.321{\pm}0.046$ \\
\midrule
\# Wins &  & $\mathbf{16}$ & $\mathbf{16}$ & 5 & 10 & 3 & 0 & 0 \\
Avg.\ Rank &  & $\mathbf{2.40}$ & 2.62 & 2.82 & 3.54 & 3.68 & 5.76 & 6.88 \\
\bottomrule
\end{tabular}%
}
\end{table}

\section{Random drop task details}
\label{app:ablations}

To study generalization to irregular sampling, we apply a custom \emph{random drop} experiment to the EigenWorms dataset: per training seed, a random fraction $r_\mathrm{train}$ of time indices is sampled uniformly without replacement and discarded, and the model observes only the remaining $(1-r_\mathrm{train})$ fraction of steps. The retained observations are paired with their original timestamps so that the model can be informed of the size of the gaps caused by the random drops. For the models that can handle time gaps, S5 and TIDES, timestep information is explicitly fed in as a learnt LTI feature only through the discretization, whereas for the Mamba family models, it is concatenated with the input and treated as an extra channel. At evaluation time, we fix the dropped indices per sequence (drawn once per seed before training begins) and sweep $r_\mathrm{test} \in \{0.1, 0.3, 0.5, 0.7, 0.9\}$ independently of the training rate. This separation between $r_\mathrm{train}$ and $r_\mathrm{test}$ turns the benchmark into an \emph{extrapolation} test: models must handle observation densities that may be substantially higher or lower than those seen during training, with $r_\mathrm{test}{=}0.9$ representing the most extreme out-of-distribution regime (only 10\% of steps observed), enforcing true continuous learning. 

Within this task, we ablate the role of input dependence in each component of the SSM by defining six architecture variants that share all hyperparameters ($L{=}1$, $P{=}16$, \texttt{bidir}, ZOH, $\mathrm{wd}{=}0.1$) and adjust the hidden dimension $d$ to keep the number of parameters close among the models, matched to ${\approx}30\mathrm{k}$ parameters. Additionally, we benchmark three official Mamba variants, namely, Mamba, Mamba-2, and Mamba-3. Each variant selects independently whether $\mathrm{Re}(\Lambda)$, $\mathrm{Im}(\Lambda)$, the $B{,}C$ projections, and the step size $\Tilde{\Delta}$ are LTI or input-dependent (see Table~\ref{tab:droprate_full}). $\Tilde{\Delta}$ being LTI means we directly provide the physical $\Delta$ to the model, whereas $\Tilde{\Delta}$ being ID assumes $\Delta$ is an additional input channel.

Models are trained with drop rate $r_\mathrm{train}{=}0.5$ on EigenWorms and evaluated across $r_\mathrm{test} \in \{0.1, 0.3, 0.5, 0.7, 0.9\}$ (Fig.~\ref{fig:droprate}).

\paragraph{LTI $B{,}C$ lacks expressivity.}
S5 and TIDES$_\Lambda$, which use LTI $B$ and $C$, plateau at a poor accuracy across $r_\mathrm{test}$: statically deciding how the input affects the state ($B$) and how the state affects the output ($C$) significantly limits the model's performance at this task. All variants with input-dependent $B{,}C$ (Mamba$_{\text{S}}$, TIDES$_\mathrm{BC}$, TIDES, TIDES$_\mathrm{full}$) learn the task successfully, confirming that the ability to expressively modulate what is read and written at each observation is essential.

\paragraph{Input-dependent $\tilde\Delta$ hurts out-of-distribution generalization.}
Mamba$_{\text{S}}$, which makes the step size $\tilde\Delta$ input-dependent instead of $\Lambda$, matches TIDES on in-distribution rates ($r = {\sim}0.5$) but collapses dramatically at $r_\mathrm{test}{=}0.9$, dropping from ${\sim}0.7$ to ${\sim}0.4$. Because $\tilde\Delta_k$ is computed as $\mathrm{softplus}(W_{\Delta}[u_k, \Delta_k])$, it cannot extrapolate to out-of-distribution drop rates.

\paragraph{Input-dependent $\Lambda$ and $B{,}C$ unlocks maximal expressivity without sacrificing extrapolation.}
TIDES, which adds input-dependent $\mathrm{Re}(\Lambda)$ on top of input-dependent $B{,}C$, achieves the best mean accuracy ($0.739$) and stays robust across drop rates, indicating that dynamic real eigenvalues help the model adapt its decay rate to varying observation densities. Making $\mathrm{Im}(\Lambda)$, which models frequency oscillation, input-dependent as well (TIDES$_\mathrm{full}$) enables per-time-step modeling of frequency oscillations, which does not hurt extrapolation; however, it slightly hurts accuracy, suggesting the imaginary component is best kept static. 

\paragraph{RFormer as a non-SSM baseline.}
To situate our SSM ablation against a qualitatively different model family, we also evaluate RFormer \citep{moreno2024rough}, a Transformer that replaces the raw time series with fixed-length path-signature summaries computed over uniformly spaced windows in the original time grid, thereby handling irregular observations without recurrence. We use the published best-configuration for EigenWorms reported in the RFormer paper (10 local windows, signature level~2, $n_\mathrm{embd}{=}20$, $n_\mathrm{head}{=}1$, $L{=}2$). RFormer achieves a mean accuracy of $0.720$ across drop rates, competitive with but below TIDES ($0.739$). More strikingly, it exhibits the same sharp collapse at $r_\mathrm{test}{=}0.9$ ($0.562$) observed for Mamba$_{\text{S}}$. The failure mode is structural: with 10 fixed windows, each window contains ${\approx}900$ observations at $r_\mathrm{train}{=}0.5$ but only ${\approx}180$ at $r_\mathrm{test}{=}0.9$, placing the signature features far outside the training distribution. Since \texttt{num\_windows} is baked into the positional embeddings and attention mask, more windows cannot be used at test time without retraining, preventing generalization on OOD drop rates. The comparison is, therefore, structurally fair, and the result underscores that TIDES's robustness at extreme drop rates stems from a genuine architectural advantage: continuous-time eigenvalue extrapolation via $e^{\Lambda \Tilde{\Delta}_k}$, which neither input-dependent $\Tilde{\Delta}$ nor fixed signature windows can match.

\begin{table}[t]
\centering
\caption{Architecture and drop-rate generalization results on EigenWorms
         ($r_\mathrm{train}{=}0.5$). Shared SSM settings: $L{=}1$, $P{=}16$,
         \texttt{bidir}, ZOH, \texttt{wd}{=}0.1, lr${=}10^{-3}$, 400 epochs.
         Accuracy is mean over 3 seeds.}
\label{tab:droprate_full}
\setlength{\tabcolsep}{4pt}
\begin{tabular}{l r cccc ccccc c}
\toprule
\multicolumn{6}{c}{\textbf{Architecture}} & \multicolumn{6}{c}{\textbf{Accuracy}} \\
\cmidrule(lr){1-6}\cmidrule(lr){7-12}
\textbf{Model} & $h$ & $\mathrm{Re}(\Lambda)$ & $\mathrm{Im}(\Lambda)$ & ${B{,}C}$ & $\Tilde{\Delta}$
  & $r{=}0.1$ & $r{=}0.3$ & $r{=}0.5$ & $r{=}0.7$ & $r{=}0.9$ & {Mean} \\
\midrule
  S5 & 80 & \textsc{lti} & \textsc{lti} & \textsc{lti} & \textsc{lti} & $0.419$ & $0.419$ & $0.419$ & $0.419$ & $0.419$ & $0.419$ \\
  Mamba$_{\text{S}}$ & 16 & \textsc{lti} & \textsc{lti} & \textsc{id} & \textsc{id} & $0.705$ & $0.705$ & $0.705$ & $0.686$ & $0.400$ & $0.640$ \\
  Mamba$^\ast$ & 16 & \textsc{lti} & \textsc{-} & \textsc{id} & \textsc{id} & $0.752$ & $\mathbf{0.771}$ & $\mathbf{0.771}$ & $\mathbf{0.762}$ & $0.419$ & $0.695$ \\
  Mamba-2$^\ast$ & 16 & \textsc{lti} & \textsc{-} & \textsc{id} & \textsc{id} & $0.743$ & $0.752$ & $0.762$ & $0.743$ & $0.457$ & $0.691$ \\
  Mamba-3$^{\ast, 1}$ & 16 & \textsc{id} & \textsc{id} & \textsc{id} & \textsc{id} & $0.733$ & $0.743$ & $0.752$ & $0.733$ & $0.562$ & $0.705$ \\
 \textbf{TIDES} & 16 & \textsc{id} & \textsc{lti} & \textsc{id} & \textsc{lti} & $0.752$ & $0.743$ & $0.743$ & $0.743$ & $\mathbf{0.714}$ & $\mathbf{0.739}$ \\
  TIDES$_\Lambda$ & 80 & \textsc{id} & \textsc{lti} & \textsc{lti} & \textsc{lti} & $0.457$ & $0.457$ & $0.457$ & $0.467$ & $0.467$ & $0.461$ \\
  TIDES$_\mathrm{BC}$ & 16 & \textsc{lti} & \textsc{lti} & \textsc{id} & \textsc{lti} & $0.695$ & $0.695$ & $0.695$ & $0.695$ & $0.686$ & $0.693$ \\
  TIDES$_\mathrm{full}$ & 16 &  \textsc{id} & \textsc{id} & \textsc{id} & \textsc{lti} & $0.676$ & $0.667$ & $0.676$ & $0.648$ & $0.648$ & $0.663$ \\
\midrule
  RFormer$^\dagger$ & $20^\dagger$ &  \multicolumn{4}{c}{\textit{path signature}} & $\mathbf{0.771}$ & $0.762$ & $\mathbf{0.771}$ & $0.733$ & $0.562$ & $0.720$ \\
\bottomrule
\end{tabular}
\vspace{4pt}
\begin{minipage}{\linewidth}
{\small 
$^\ast$~Weight-decay set to 0 for the Mamba family. \\
$^{1}$~Mamba-3 uses trapezoidal discretization, instead of ZOH. \\
$^\dagger$~RFormer uses the best published EigenWorms configuration from \citet{moreno2024rough} (10 local path-signature windows, level~2, $n_\mathrm{embd}{=}20$, $n_\mathrm{head}{=}1$,
lr${=}6.73{\times}10^{-3}$, 400 epochs with best validation accuracy model selection).
}
\end{minipage}
\end{table}

\section{Training time and memory}
\label{app:resources}
SSMs scale linearly in sequence length, $\mathcal{O}(L)$, while attention-based transformers incur quadratic cost, $\mathcal{O}(L^2)$, in both compute and memory. This asymptotic gap is reflected in Table~\ref{tab:time_memory}: at short sequences ($L=1{,}000$), RFormer is faster and more memory-efficient than TIDES due to smaller constant factors, but the crossover occurs well before $L=5{,}000$, where TIDES is already $2.2\times$ faster and uses roughly half the memory. By $L=10{,}000$, RFormer exhausts the 24\,GB budget entirely, while TIDES continues to scale linearly in both wall-clock time and peak memory. This confirms that the linear-recurrence formulation of TIDES retains the favorable scaling characteristics of the SSM family and makes it suitable for long sequences.

\begin{table}[H]
\centering
\caption{Training step wall time (ms, mean $\pm$ std over 20 steps) and peak GPU memory
         for TIDES and RFormer across sequence lengths $L$ (batch size 8, channels 1).
         Both models are parameter-matched at ${\approx}100$k trainable parameters with 4 layers.
         OOM denotes out-of-memory on a 24\,GB Quadro RTX 6000 GPU.}
\label{tab:time_memory}
\setlength{\tabcolsep}{6pt}
\begin{tabular}{r rr rr}
\toprule
 & \multicolumn{2}{c}{\textbf{TIDES}} & \multicolumn{2}{c}{\textbf{RFormer}} \\
\cmidrule(lr){2-3} \cmidrule(lr){4-5}
$L$ & Time (ms) & Mem (MB) & Time (ms) & Mem (MB) \\
\midrule
1\,000   & $59.9 \pm 1.3$ & 1\,223  & $22.3 \pm 1.6$ & 531     \\
5\,000   & $173.3 \pm 0.1$ & 5\,924  & $380.1 \pm 0.4$ & 11\,508 \\
10\,000  & $342.1 \pm 0.3$ & 11\,816 & \multicolumn{2}{c}{OOM}  \\
\bottomrule
\end{tabular}
\end{table}

\newpage
\section*{NeurIPS Paper Checklist}

\begin{enumerate}

\item {\bf Claims}
    \item[] Question: Do the main claims made in the abstract and introduction accurately reflect the paper's contributions and scope?
    \item[] Answer: \answerYes{}    
    \item[] Justification: The abstract and Section~\ref{sec:intro} state three claims --- (i) routing input-dependence through $\mathrm{Re}(\Lambda)$ and $B,C$ rather than $\Delta$ preserves both per-token expressivity and a physical sampling interval; (ii) a controlled diagnostic (Fading Flash) isolates three failure modes of existing SSMs; (iii) TIDES achieves new state-of-the-art on UEA time-series classification and the Physiome-ODE regression benchmark. Each is substantiated by the corresponding section, respectively, Section~\ref{sec:method}, Section~\ref{sec:toy}; and  Section~\ref{sec:experiments} (Tables~\ref{tab:uea}, \ref{tab:results}).

\item {\bf Limitations}
    \item[] Question: Does the paper discuss the limitations of the work performed by the authors?
    \item[] Answer: \answerYes{}  
    \item[] Justification: Section \ref{sec:discussion} (Limitations) flags the parameter and compute overhead of the input-dependent projectors relative to S5, the per-step projection bottleneck for very long sequences ($L > 10^5$), and the restriction of our evaluation to time-series benchmarks (no language or other modalities without a clear physical sampling interval). Furthermore, Appendix \ref{app:resources} discusses the computational efficiency of TIDES.
    

\item {\bf Theory assumptions and proofs}
    \item[] Question: For each theoretical result, does the paper provide the full set of assumptions and a complete (and correct) proof?
    \item[] Answer: \answerNA{}  
    \item[] Justification: The paper does not include formal theorems or proofs. The closed-form ZOH discretization expressions used in Section \ref{sec:method} and Appendix \ref{app:disc} are standard textbook derivations from the linear-system literature \citep{smith2023s5,gu2023mamba} and are merely stated, not proved as new results. The contributions are architectural, not theoretical.  

    \item {\bf Experimental result reproducibility}
    \item[] Question: Does the paper fully disclose all the information needed to reproduce the main experimental results of the paper to the extent that it affects the main claims and/or conclusions of the paper (regardless of whether the code and data are provided or not)?
    \item[] Answer: \answerYes{}      
    \item[] Justification: The TIDES architecture is fully specified in Section \ref{sec:method} and Appendix \ref{app:architecture}, including the input-dependent projector parameterization, low-rank factorization of $B,C$, HiPPO bias initialization, ZOH discretization (Appendix \ref{app:disc}), and per-block components (BatchNorm, GLU, residual). The Fading Flash experiment --- data generation, model variants, training protocol --- is described in Section \ref{sec:toy} and Appendix \ref{app:toy}. For UEA (Appendix \ref{app:uea}), we follow the LogNCDE evaluation protocol bit-for-bit (six fixed datasets, $70/15/15$ random splits across five seeds, per-channel train-statistics standardization). For Physiome-ODE (Appendix \ref{app:physiome-data}) we use the available public benchmark \cite{klotergens2025physiome} without modification, including their $5$-fold cross-validation, $80\%$ random masking, and the $10$-config hyperparameter search on the first fold. Per-dataset hyperparameter ranges and random seeds are listed in Appendix \ref{app:experiments}. The Supplemental Materials include the code and the data required to reproduce the reported results.

\item {\bf Open access to data and code}
    \item[] Question: Does the paper provide open access to the data and code, with sufficient instructions to faithfully reproduce the main experimental results, as described in supplemental material?
    \item[] Answer: \answerYes{} 
    \item[] Justification:  An anonymized code snapshot is included as supplementary material (\texttt{tides.zip}); its \texttt{README.md} lists environment setup, and the exact commands to reproduce Tables \ref{tab:uea} and \ref{tab:results} and the Fading Flash diagnostic. A non-anonymized release will follow at camera-ready.  

\item {\bf Experimental setting/details}
    \item[] Question: Does the paper specify all the training and test details (e.g., data splits, hyperparameters, how they were chosen, type of optimizer) necessary to understand the results?
    \item[] Answer: \answerYes{} 
    \item[] Justification: Section \ref{sec:experiments} states the optimizer, schedule, and loss; Appendix \ref{app:experiments} lists per-benchmark splits (UEA: $70/15/15$ random partition over five seeds; Physiome-ODE: the benchmark's published five-fold CV with $80\%$ masking), hyperparameter search ranges, and the selection protocol (10 random configurations chosen on the first fold's validation split, then retrained on all folds \cite{klotergens2025physiome}).

\item {\bf Experiment statistical significance}
    \item[] Question: Does the paper report error bars suitably and correctly defined or other appropriate information about the statistical significance of the experiments?
    \item[] Answer: \answerYes{} 
    \item[] Justification: Every entry in the UEA table (Table \ref{tab:uea}) and the Physiome-ODE table (Table \ref{tab:results}) is reported as mean $\pm$ one standard deviation across five independent runs. One standard deviation is selected over 2 standard deviations to ensure a fair comparison with the baselines reported in \cite{klotergens2025physiome} and \cite{moreno2024rough}. Aggregate columns (Avg., Avg.\ Rank, \# Wins) are computed deterministically from the fold-mean entries. Standard deviations are reported (not standard errors of the mean) and were computed with the unbiased estimator (\texttt{numpy.std} with \texttt{ddof=1}); we make no Normality assumption and do not perform pairwise hypothesis tests, since the across-method ranking is already separated by several standard deviations on most datasets and the average rank statistic is order-based.

\item {\bf Experiments compute resources}
    \item[] Question: For each experiment, does the paper provide sufficient information on the computer resources (type of compute workers, memory, time of execution) needed to reproduce the experiments?
    \item[] Answer: \answerYes{} 
    \item[] Justification: Section \ref{sec:experiments} reports approximately 5,000 GPU-hours of A100 time. Appendix \ref{app:experiments} further reports per-run wall time and peak memory across sequence lengths (Table \ref{tab:time_memory}).
    
\item {\bf Code of ethics}
    \item[] Question: Does the research conducted in the paper conform, in every respect, with the NeurIPS Code of Ethics \url{https://neurips.cc/public/EthicsGuidelines}?
    \item[] Answer: \answerYes{} 
    \item[] Justification: The research conducted confirms, in every respect, with the NeurIPS Code of Ethics.

\item {\bf Broader impacts}
    \item[] Question: Does the paper discuss both potential positive societal impacts and negative societal impacts of the work performed?
    \item[] Answer: \answerNA{}  
    \item[] Justification: There is no direct societal impact of the work performed.
    
\item {\bf Safeguards}
    \item[] Question: Does the paper describe safeguards that have been put in place for responsible release of data or models that have a high risk for misuse (e.g., pre-trained language models, image generators, or scraped datasets)?
    \item[] Answer: \answerNA{} 
    \item[] Justification: The paper poses no such risks.

\item {\bf Licenses for existing assets}
    \item[] Question: Are the creators or original owners of assets (e.g., code, data, models), used in the paper, properly credited and are the license and terms of use explicitly mentioned and properly respected?
    \item[] Answer: \answerYes{} 
    \item[] Justification:  All existing assets are cited at first use. The UEA archive \citep{bagnall2018uea} is publicly distributed for research; the Physiome-ODE benchmark \citep{klotergens2025physiome} is released by the    authors under MIT. Baseline implementations are taken unmodified from the authors' public repositories under their MIT or Apache-2.0 licenses.

\item {\bf New assets}
    \item[] Question: Are new assets introduced in the paper well documented and is the documentation provided alongside the assets?
    \item[] Answer: \answerYes{} 
    \item[] Justification: The paper introduces (i) a PyTorch implementation of the TIDES architecture and the training pipeline used for UEA and Physiome-ODE, and (ii) the Fading Flash diagnostic --- data generation, model variants, and training protocol --- defined in Section \ref{sec:toy} and Appendix \ref{app:toy}. Both will be released with the camera-ready version under an open-source license. No new dataset is introduced; UEA and Physiome-ODE are used as published.

\item {\bf Crowdsourcing and research with human subjects}
    \item[] Question: For crowdsourcing experiments and research with human subjects, does the paper include the full text of instructions given to participants and screenshots, if applicable, as well as details about compensation (if any)? 
    \item[] Answer: \answerNA{} 
    \item[] Justification: The paper does not involve crowdsourcing nor research with human subjects.

\item {\bf Institutional review board (IRB) approvals or equivalent for research with human subjects}
    \item[] Question: Does the paper describe potential risks incurred by study participants, whether such risks were disclosed to the subjects, and whether Institutional Review Board (IRB) approvals (or an equivalent approval/review based on the requirements of your country or institution) were obtained?
    \item[] Answer: \answerNA{}  
    \item[] Justification: The paper does not involve crowdsourcing nor research with human subjects.

\item {\bf Declaration of LLM usage}
    \item[] Question: Does the paper describe the usage of LLMs if it is an important, original, or non-standard component of the core methods in this research? Note that if the LLM is used only for writing, editing, or formatting purposes and does \emph{not} impact the core methodology, scientific rigor, or originality of the research, declaration is not required.
    \item[] Answer: \answerNA{} 
    \item[] Justification: The core method development in this research does not involve LLMs as any important, original, or non-standard components.

\end{enumerate}

\end{document}